\documentclass{article}

\usepackage{arxiv}

\usepackage[utf8]{inputenc} 
\usepackage[T1]{fontenc}    
\usepackage{hyperref}       
\usepackage{url}            
\usepackage{booktabs}       
\usepackage{amsfonts}       
\usepackage{nicefrac}       
\usepackage{microtype}      
\usepackage{lipsum}
\usepackage{graphicx}
\usepackage{amsmath,amsfonts}
\usepackage{algorithmic}
\usepackage{algorithm}
\usepackage{array}
\usepackage[caption=false,font=normalsize,labelfont=sf,textfont=sf]{subfig}
\usepackage{textcomp}
\usepackage{stfloats}
\usepackage{url}
\usepackage{verbatim}
\usepackage{graphicx}
\usepackage{cite}
\usepackage{xcolor}
\usepackage{amsmath}
\usepackage{makecell}
\usepackage{multicol}
\usepackage{multirow}
\usepackage{booktabs}
\usepackage{siunitx}
\usepackage{gensymb}
\usepackage{soul, color, xcolor}
\usepackage[utf8]{inputenc}
\usepackage[english]{babel}
\usepackage{threeparttable}
\usepackage{graphicx} 
\usepackage{amsmath}
\usepackage{amssymb}
\usepackage{tablefootnote}
\usepackage{stfloats}
\usepackage{threeparttable}
\usepackage{color,soul}
\graphicspath{ {./images/} }

\title{Compliant actuators that mimic biological muscle performance with applications in a highly biomimetic robotic arm}

\author{
Haosen Yang \\
  The Department of Mechanical, Aerospace and Civil Engineering\\
  University of Manchester\\
  Manchester, M13 9PL, UK \\
  \texttt{haosen.yang@postgrad.manchester.ac.uk} \\
   \And
 Guowu Wei \\
  School of Science, Engineering and Environment\\
  University of Salford\\
  Salford, M5 4WT, UK \\
  \texttt{g.wei@salford.ac.uk} \\
  \And
Lei Ren \\
  The Key Laboratory of Bionic Engineering, Ministry of Education\\
  Jilin University\\
  Changchun 130025, China \\
  \texttt{lren@jlu.edu.cn} \\
\And
Lingyun Yan\\
Department of Robotics Engineering, School of Electrical and Electronic Engineering\\
Shanghai Institute of Technology\\
Shanghai, 200235, China\\
  \texttt{lingyunyan@sit.edu.cn}
}
\begin{document}
\maketitle
\begin{abstract}

This paper endeavours to bridge the existing gap in muscular actuator design for ligament-skeletal-inspired robots, thereby fostering the evolution of these robotic systems. We introduce two novel compliant actuators, namely the Internal Torsion Spring Compliant Actuator (ICA) and the External Spring Compliant Actuator (ECA), and present a comparative analysis against the previously conceived Magnet Integrated Soft Actuator (MISA) through computational and experimental results. These actuators, employing a motor-tendon system, emulate biological muscle-like forms, enhancing artificial muscle technology. A robotic arm application inspired by the skeletal ligament system is presented. Experiments demonstrate satisfactory power in tasks like lifting dumbbells (peak power: 36W), playing table tennis (end-effector speed: 3.2 m/s), and door opening, without compromising biomimetic aesthetics. Compared to other linear stiffness serial elastic actuators (SEAs), ECA and ICA exhibit high power-to-volume (361 x 10³ W/m³) and power-to-mass (111.6 W/kg) ratios respectively, endorsing the biomimetic design's promise in robotic development.

\end{abstract}


\maketitle

\section{Introduction}

The escalating integration of robots and humans in intricate environments accentuates the imperative of flexibility and safety in design. Two prevalent methodologies, algorithmic and mechanical compliance, are employed to fortify safety in robotic systems. Algorithmic compliance maintains rigidity in robots while modulating joint torque\cite{hyon2007full}, albeit susceptible to sensor data inaccuracies, necessitating recalibration in divergent environments. Conversely, mechanical compliance enhances safety by absorbing unforeseen impacts, even sans power\cite{liu2021whole}. Pratt et al.\cite{pratt1995proceedings} made the first proposal for a Series Elastic Actuator (SEA) to achieve mechanical compliance through the utilization of an elastic element that provides compliant coupling between the geared motor and the end effector\cite{liu2021whole}, followed by the advent of Variable Stiffness Actuators (VSA) that mimic biological muscle attributes, finding utility in soft and exoskeleton robotics\cite{schrade2018development,shao2021design,xu2021design,hussain2021design,li2020design}. These technological incorporations propel robotic systems towards enhanced safety and flexibility.

While rotational SEA and VSA discussed above which are based on the hinge-joint design scheme may be effective for providing compliance in industrial and {traditional robots}\cite{englsberger2014overview, yu2014design, paik2012development, tsagarakis2011design, colasanto2012compact, tsagarakis2017walk}, they are not well-suited in the development of a highly biomimetic robot. Tendon-driven actuation, essential for enhanced biomimicry, desires an actuator emulating biological muscle. This technique endows actuators with muscle-like appearance and functionality\cite{sodeyama2008designs, marques2010ecce1, asano2019musculoskeletal}. Tendon-driven robots employ electronic motors, positioned either remotely or locally\cite{toshimitsu2021biomimetic, fan2022feedforward}. However, in most tendon-driven robotic limb and assistive exoskeletons designs combined with SEAs or VSAs\cite{hyun2020singular, lee2022development, herbin2021human, zhu2021design, nakanishi2011development, kobayashi2010development, mouthuy2022humanoid, kawaharazuka2022robust}, the motors are positioned remotely to allow the use of large motors. This configuration may cause considerable frictional challenges, and decrease the precision of joint movements and the effectiveness of joint torque. A more desirable design is a local tendon-driven scheme combined with either SEAs or VSAs. By utilizing this approach, the actuator can be situated closer to the joint, which allows for better emulation of the properties of biological muscles and avoids the interactions caused by tendons crossing multiple joints. This approach can lead to improved performance and more natural movements for robotic systems designed to mimic human motion.

This paper presents two innovative compact motor-based, muscle-like actuators employing a local tendon-driven mechanism and benefiting from SEA designs. The elastic elements in these actuators differ—being linear as seen in compression and torsion springs. Each actuator exhibits advantages and limitations regarding compactness and speed. A comparison is drawn with a prior design, MISA\cite{yang2023novel}, which utilizes non-linear stiffness elastic elements like oppositely placed magnets. The suitability of each actuator varies across application scenarios, and a thorough analysis delineates the trade-offs, guiding researchers and engineers in choosing the right actuator for their applications, thereby informing future robotic design endeavours.

\section{Related work}

\subsection{Motor-based compliant actuators}

The Hill-based model from the 1930s is a widely accepted representation of human muscle\cite{hill1938heat}. It consists of three key components: the parallel element (PE), the series element (SE), and the contractile element (CE). The PE represents the mechanical properties of the muscle fibre membrane, muscle fascia and other connective tissue when the muscle is relaxed. The SE represents the inherent elasticity of the muscle fibres, and the CE represents the muscle fibres themselves. Based on this model, researchers have proposed SEAs which add a series elements with constant elasticity and damping to decouple the motor (CE) from the end effector, so that the output force of the SEA corresponds to its deformation. {SEAs are widely used in a variety of applications, including exoskeleton robots} \cite{hussain2021exoskeleton, zhang2019admittance, chen2019elbow, herbin2021human, marconi2019novel, zhong2021toward}, lower limb robots \cite{lee2019development, hong2020combined, zhu2016adaptive, accoto2013design, pratt2004series}, prosthetics \cite{sun2018variable, convens2019modeling, rouse2014clutchable}, humanoid robots \cite{knabe2015design}, {rehabilitation robots}\cite{yu2015human, yu2013novel} and other areas. 

The most conventional SEAs are rotary, linear, and tendon-driven. Rotary SEAs add linear stiffness elastic elements, such as torsion springs, between the motors and the driven joints \cite{kong2009control, ragonesi2011series}. These actuators are typically placed at the joint and drive the joint directly, which may increase the size of the robot joint. Linear SEAs add extension springs between linear actuators and the driven arms \cite{paine2012new, truong2020design} which are usually longer in length. For SEAs combined with tendon-driven methods, series extension springs \cite{rouse2014clutchable} are widely used, or torsion springs are added between motors and the driving pulleys\cite{lu2015design}. However, these SEAs are usually used for remote tendon-driven joints, which are not designed to be muscle-shaped and therefore difficult to implement in highly biomimetic robots.





Magid \& Law investigated the length-tension relationship of the frog muscle \cite{magid1985myofibrils} in 1895. It shows the non-linear property, where the curve slope increases as the length increases. To emulate the biologic muscle properties, VSA was developed. A Non-linear elastic element or a stiffness adjustable device were used to represent the SEs in the Hill-based model \cite{tsagarakis2009compact, wolf2011dlr, guo2016mechanical, thorson2011nonlinear, wolf2008new, tonietti2005design, schiavi2008vsa, palli2011design}. Similar to SEAs, VSAs can be classified as rotary and tendon-driven types. Typical rotary VSAs use linear springs sets\cite{nakanishi2011development, thorson2011nonlinear, qian2022design, toubar2022design}, elastomer-based material \cite{abe2012control}, or non-linear stiffness torsion springs\cite{yildirim2021design, accoto2013design, chen2017design}. The joints are driven by the rotary VSAs directly, with the elastic elements placed between the driven joints and the motors. For VSAs with tendon-driven joints, spring-pulley systems are normally used \cite{chen2019elbow, mori2019high, li2020design}, which allow the VSAs to be capable of achieving an infinite range of joint stiffness. Another commonly used mechanism is connecting tendons to the driven joint via a non-linear spring \cite{kobayashi2010development}. Compared to linear stiffness SEAs, robot joints equipped with VSAs are able to present softer joints when receiving a small external force. When the impact force dramatically increases, the joint stiffness increases accordingly. 
 
Most current SEAs and VSAs are intended for applications where the joints are driven directly by motors or remotely by motors through the tendon. For rotary compliant actuators {at joints where the motor shaft and joint rotation axis coincide}, the integration of motors, gearboxes, and soft elements at the joints usually results in oversized joints compared to biological joints. For current SEAs and VSAs integrated tendon-driven schemes, due to the use of remote-driven design, these actuators require large support systems including motor support and tendon guidance systems. They abandoned muscle-like shapes, making it difficult to achieve a human-like form in the serving robots. In addition, the tendon-driven method can be challenged by the requirement to drive a joint through multiple joints. For example, to drive the elbow joint, a motor placed in the torso may require a tendon passing through the shoulder joint, resulting in interactions between joint actuation. Additionally, the employment of extended tendons may cause considerable frictional challenges. The stretching of tendons and their ensuing durability constitute concerns, thereby decreasing the precision of joint movements. A guiding tube is required to direct the tendon to the driven joint. This may limit the tensile force transmitted by the tendon, thereby restricting the joint's torque output. 

Optimizing motor-based compliant actuators for local tendon-driven artificial muscles with muscle-like shapes and properties is an innovative design direction. These actuators must be compact, lightweight, and possess high performance, including a high power-to-volume ratio, high power-to-mass ratio, large elastic travel, and a wide range of stiffness variation for VSA. However, designing motor-based artificial muscles that can be applied to various joints without modification is challenging. Unlike the remote control scheme that allows for large-size motors and complicated soft components, this new design direction requires minimizing size and weight while maintaining a muscle-like shape. Therefore, it remains to be determined whether actuators based on this design concept can drive highly biomimetic biological robots with high output capacity. Further research and development in this direction may be necessary to address these challenges and explore the potential for local tendon-driven artificial muscles to drive highly biomimetic robots.

\subsection{Biomimetic robots}

For decades, robots have predominantly utilized geared motors for joint actuation, directly connecting motors to hinge or universal joints. Notable instances of such designs encompass the Robotic arm and hand system \cite{paik2012development}, BHR-5 robot \cite{yu2014design}, ASIMO robot\cite{sakagami2002intelligent}, WABIAN-2 robot \cite{ogura2006development}, THBIP-2 robot\cite{xia2008design}, Phoenix robot\cite{nichols2023artificial}, and Tesla robot\cite{ackerman2022robotics}. These systems, while offering merits like streamlined design, ease of maintenance, cost-efficiency, and precise movement, diverge fundamentally from human biomechanics. Key limitations include the vulnerability of high-torque motors and their reduction gear systems to sudden impacts, risking damage. Such vulnerabilities raise safety concerns, especially in human-robot interaction contexts. Additionally, the cascading of single-DOF joints to achieve multi-DOF configurations often necessitates increased spatial allocations, thereby constraining operational versatility.

A paradigm shift towards highly biomimetic robots is evident in contemporary robotics. These robots emulate human musculoskeletal architectures, encompassing bones, joints, and musculature. They leverage tendon-driven mechanisms, inspired by human joints, to foster human-analogous kinematics and aesthetics. Pioneering in this arena are models like the Kenshiro robot\cite{asano2019musculoskeletal}, Kojiro robot\cite{sodeyama2008designs}, ECCE\cite{marques2010ecce1}, and Roboy\cite{mouthuy2022humanoid} which may employ ligaments for joint stabilization. However, a critique emerges in their often-overly simplified replication of human anatomy. While striving for functional equivalence, they sometimes neglect essential soft components. Instead of artificial muscles, many employ motors positioned remotely, actuating joints via tendons, which might introduce disparities in form and size relative to human counterparts. Central to advancing such robots is the evolution of artificial muscle designs. Researchers have innovated diverse artificial muscle variants, including pneumatic muscles\cite{mori2019high} and twisted and coiled fibers\cite{haines2014artificial}. These actuators, characterized by inherent compliance, can be tailored for specific applications. Nonetheless, challenges persist: some variants, which depend on thermal contractions, confront issues with precise control and rapid cooling, making motor-based soft artificial muscles a prevalent choice.

This paper seeks to bridge the existing gap, presenting a comparative analysis of three motor-based muscle-like actuators, amalgamating local-tendon-driven modalities with SEA and VSA designs.

\section{Design and characteristics of compliant actuators}

\begin{figure*}[htbp]
\centerline{\includegraphics[width=1\textwidth]{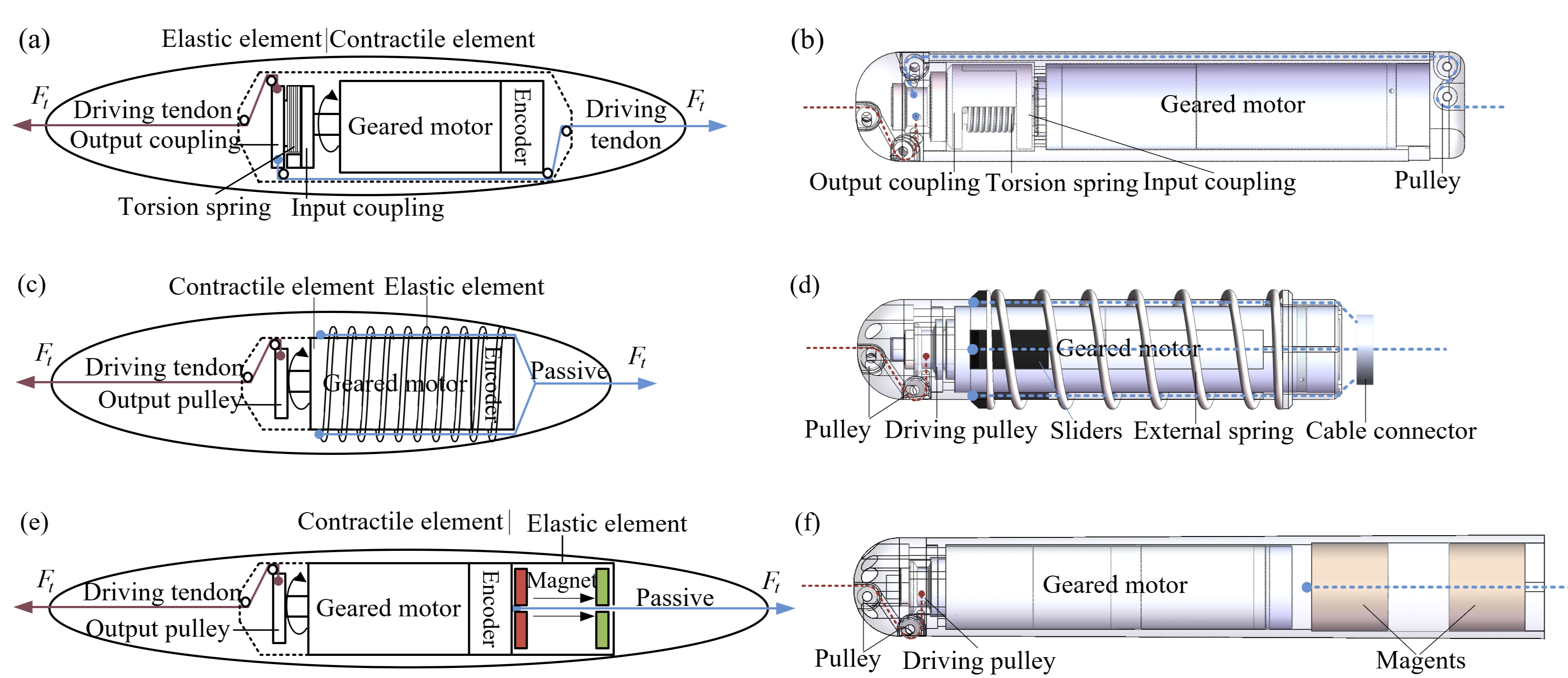}}
\caption{Schematic diagrams and CAD models of three compliant actuators, including (a) (b) Internal Linear Torsion Spring Compliant Actuator (ICA); (c) (d) External Linear Spring Coft Actuator (ECA); (e) (f) Magnet Integrated Non-linear Soft Actuator (MISA)\cite{yang2023novel}.}
\label{fig3.1}
\end{figure*}

To address the problem listed above, this section introduces the design of two tendon-driven compliant actuators that can achieve various degrees of bionic properties. Fig. \ref{fig3.1} shows the design schematic diagrams and CAD models of these actuators including the Internal Linear Spring Compliant Actuator (ICA) (Fig. \ref{fig3.1}(a) and (b)), the External Linear Spring Compliant Actuator (ECA) (Fig. \ref{fig3.1}(c) and (d)), and the previous introduced Magnets Integrated Non-linear Compliant Actuator (MISA) (Fig. \ref{fig3.1}(e) and (f))\cite{yang2023novel}. For a comprehensive elucidation on the design of the MISA, we direct the reader to our preceding work.

\subsection{Internal Spring Compliant Actuator (ICA)}

In biomimetic robots, artificial muscles typically adopt an elongated configuration, as exemplified by pneumatic and hydraulic muscles. When designing artificial muscles using brushless motors, the rated power of the motor is intrinsically linked to its size. This poses a challenge: integrating a high-power motor while maintaining compact dimensions is pivotal, especially for upper limb robots. The actuators presented herein are designed with this constraint in mind, prioritizing compactness by limiting the outer diameter. The inaugural actuator ensures an outer diameter only 13.6\% greater than the motor diameter. Compliance is ingeniously introduced by interposing a torsion spring between the motor and output pulley.

Fig. \ref{fig3.1}(a) and (d) depict the ICA, which encompasses a geared motor equipped with an encoder, input and output couplings, a linear stiffness torsion spring, and micro pulleys interconnected by tendons, all housed within an elongated shell.

The ICA supports both unidirectional and bidirectional contraction modes. In bidirectional mode, tendons (color-coded in red and blue) are affixed to a driven pulley connected to the output coupling. To minimize friction, the left (red) tendon is routed through two pulleys on the left, while the right (blue) tendon traverses internal rails, emerging on the right via three pulleys. In unidirectional mode, only the left tendon is operational.

The ICA's elastic component integrates a torsion spring, flanked by input and output couplings. The motor's output shaft hosts the input coupling. The torsion spring introduces a decoupling between these couplings, ensuring that any resistance experienced by the tendon amplifies the angular disparity between the couplings, causing the torsion spring to deform. This deformation facilitates energy absorption and eventual reversion to the original state, granting the desired compliance.

The ICA presents multiple merits. Notably, its reduced outer diameter facilitates the selection of a torsion spring harmonized with the motor's diameter, ensuring protection without necessitating increase in actuator diameter. Furthermore, by leveraging active tendons at both extremities, the ICA can achieve double the output speed compared to single-ended actuators of a similar motor and driving pulley composition.

Nonetheless, challenges persist. The motor's diameter inherently delimits the dimensions of the feasible torsion spring, imposing restrictions on the attainable elasticity coefficient and peak torque. Such constraints may potentially result in a limited rated torque for the chosen motor. Opting for a diminished elasticity coefficient might cause the torsion spring to surpass its maximal deformation threshold before the motor reaches its rated torque, jeopardizing the actuator's intrinsic 'soft' characteristic.

\subsection{External Spring Compliant Actuator (ECA)}

To address the ICA's constraints concerning output force, the second actuator, ECA, prioritizes enhancing the elastic force. By increasing the actuator's outer diameter and incorporating a high stiffness compression spring around the motor, the ECA simultaneously achieves a more compact length and superior controllable output force.

Fig. \ref{fig3.1}(b) and (e) illustrate the ECA's architecture, which encompasses a geared motor (designated as the contractile element), driving and auxiliary pulleys, an enveloping external spring (termed the elastic element), tendons, and a cable connector. The contractile and elastic elements are sequentially linked, encased within a slim shell which is encircled by the external spring.

Contraction is facilitated by attaching one extremity of the left tendon to the driving pulley. This tendon then traverses two pulleys — reducing friction — and is anchored to an external attachment point. Rotation of the motor shaft prompts the driving pulley to contract this tendon.

Unique to the ECA is its compression spring situated externally to the shell. This spring's right extremity is anchored by a tab, with its left counterpart interacting with three tab-fitted sliders. These sliders, which navigate along rails on the shell, modulate the spring's deformation. Three tendons (highlighted in blue) commence within this external spring, each affixed to a slider, and converge at a mobile cable connector, not rigidly attached to the actuator. When tension is exerted on the consolidated tendon, the rightward (blue) tendon activates the cable connector, prompting the sliders and compressing the spring.

In juxtaposition with the ICA, the ECA boasts a reduced length, compensating with a larger outer diameter. This strategic envelopment of the compression spring around the motor capitalizes on the motor's inherent length, enhancing compactness — a crucial feature for applications in space-restricted joints like the deltoid muscles. By permitting a more generous outer diameter, the ECA can adapt to a spectrum of compression springs, varying in elasticity coefficients and compression thresholds, facilitating a substantial controllable output force. The ability to utilize a longer spring, without extending actuator length, offers expansive elastic displacement.

Nonetheless, the ECA's enlarged outer diameter is its conspicuous drawback, potentially resulting in disproportioned limbs. Furthermore, the inherent weight of sizable compression springs might diminish the power density.

\section{Modeling and analysis of the actuators}

\begin{figure*}[htbp]
\centerline{\includegraphics[width=1\textwidth]{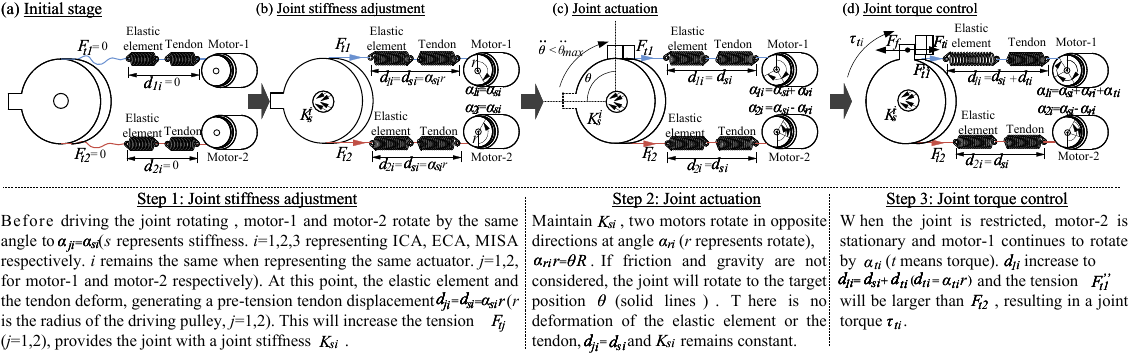}}
\caption{The process of using two compliant actuators to drive the joint movement through tendons. (a) In the initial stage, the tendons in both actuator 1 and 2 are not under tension; (b) applying joint stiffness to the joint by tensioning the tendons; (c) driving the joint to the target angle; (d) controlling the joint torque when the joint is restricted.} 
\label{fig3.2}
\end{figure*}

In tendon-driven joints, actuators in an antagonistic pair configuration are used to achieve the desired joint position $\theta$ and output torque. The three actuators (ICA, ECA, MISA) presented are appropriate for these robot joints. Fig. \ref{fig3.2} illustrates the simplified configuration of the actuation system and its corresponding operation process. The three steps for joint actuation are described. The diagram depicts the elastic elements and tendons as springs for simplification purposes.

\subsection{Force and tendon displacement relationship}

When the two ends of the tendons are limited, a tendon displacement ${d}_{i}$ (the subscript $i=1,2$ stands for ICA, ECA, respectively. The same rule is followed for other variables that contain subscript $i$) is generated due to the deformation of the elastic element and tendon when the motor rotates by angle ${\alpha}$. Assuming ${d}_{i}={d}_{mi}$ (the maximum value of $d_{i}$, the tendon deformation is included) when the elastic elements reach the limited position while the tension is ${F}_{tmi}$.

Before the elastic elements reach the limited position ($0 \leq {d}_{i} \leq {d}_{mi}$), the elastic elements and tendons are deformed. The elastic coefficient of tendon displacement in ICA due to the deformation of the torsion spring is equivalent to $k_{ts}=k_{e2}$/$2\pi r^{2}$, where $k_{e2}$ represents the torsion spring's elasticity coefficient and $r$ denotes the radius of the output pulley. In the design of ICA, the tendon passes over there supporting pulleys, leading to a friction force of ${F}_{f}=\mu_{p} {F}_{t}$, where $\mu_{p}$ represents the friction coefficient. The stiffness of the elastic element in ECA is denoted as $k_{cs}$ and corresponds to the elasticity coefficient of the compression spring. The friction can be ignored. 

The relationship function between ${F}_{t}$ and ${d}_{i}$ at this stage ${F}_{t}=f_{d} ({d}_{i})$ can be deduced as:
\begin{equation}
{d}_{i}=\begin{cases} ({F}_{t}-{F}_{f})\frac{1}{k_{ts}}+{F}_{t}\frac{1}{k_{ti}}  &i=1\\
{F}_{t}\frac{1}{k_{cs}} +{F}_{t}\frac{1}{k_{ti}} &i=2
\end{cases}
\label{eq3.1}
\end{equation}

Where $k_{ti}$ denotes the elasticity coefficient of the tendon.

After the elastic elements reach the limited position (${d}_{i}$ \textgreater  ${d}_{mi}$), ${F}_{t}$ will only stretch the tendon. ${F}_{t}=f_{d} ({d}_{i})$ in this stage is:
\begin{equation}
d_{i}=d_{mi}+({F}_{t}-{F}_{tmi})\frac{1}{k_{ti}}\label{eq3.2}
\end{equation}

By combining the two stages, the relationship between the applied force ${F}_{t}$ and the displacement $d_{i}$ can be expressed as a function:
\begin{equation}
{F}_{t}=f_{di} (d_{i})\label{eq3.3}
\end{equation}

\subsection{Joint stiffness adjustment}

\begin{figure}[htbp]
\centerline{\includegraphics[width=0.63\textwidth]{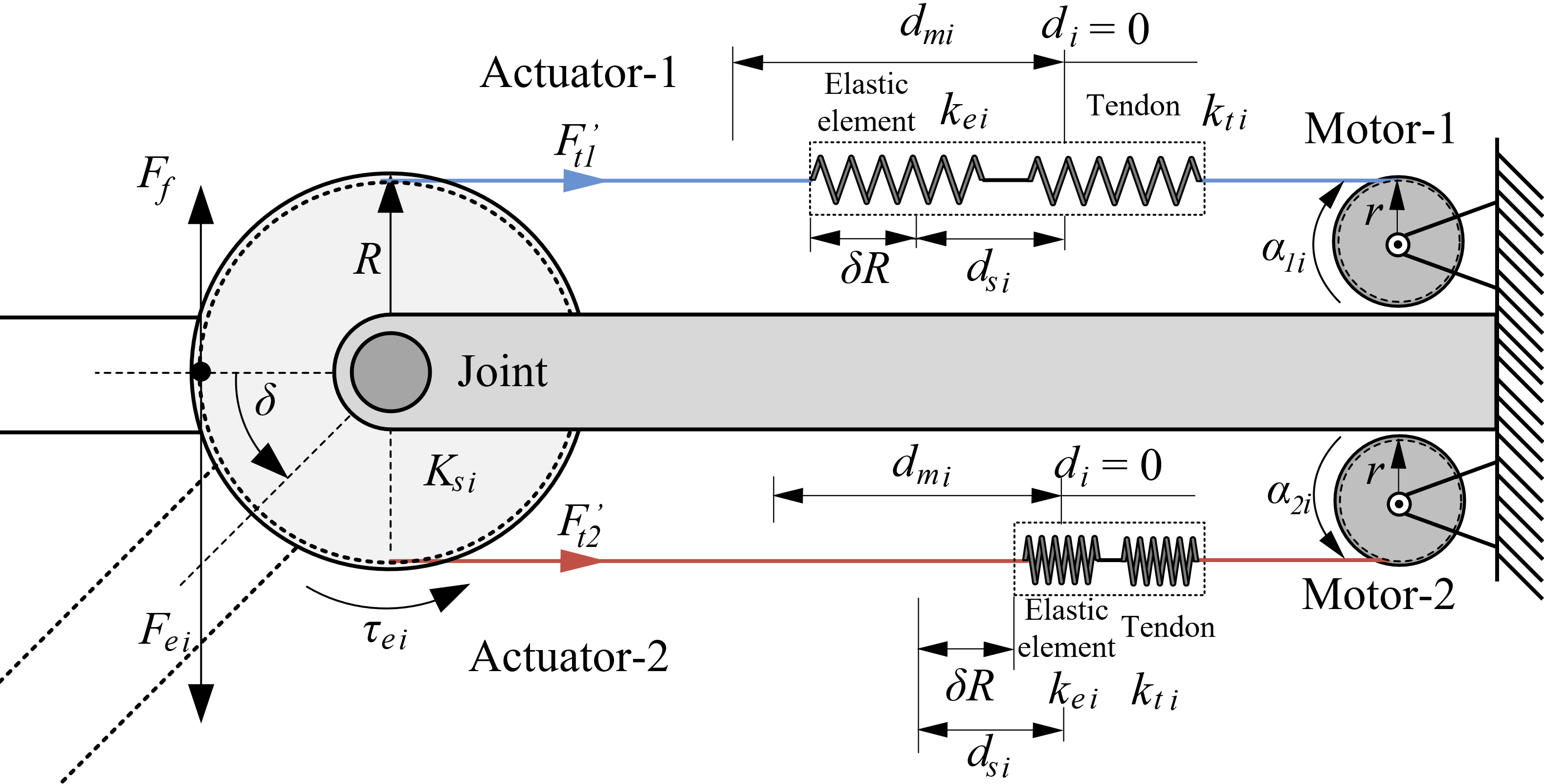}}
\caption{Joint stiffness adjustment using a pair of compliant actuators.}
\label{fig3.3}
\end{figure}

For a comprehensive comparison of ICA, ECA, and MISA, the previously established methodology\cite{yang2023novel} for joint stiffness application was reiterated. As in step 1 in Fig. \ref{fig3.2}(b), the motor 1 and 2 rotate simultaneously before actuating the joint rotates, deforming the elastic elements and tendons to produce pre-tension tendon displacement $d_{si}$ ($i=1,2$). According to the function \eqref{eq3.3}, the tension on the actuators ${F}_{tj}$ ($j=1,2$ denotes the actuator 1 and actuator 2 respectively. The same rule is followed for other variables that contain subscript $j$) is:
\begin{equation}
{F}_{tj}=f_{di}(d_{si})  \&(i=1,2; j=1,2)\label{eq3.4}
\end{equation}

A joint stiffness $K_{si}$ will be applied to the joint. A small $K_{si}$ may lead the joint to oscillate easily under system vibration and a large $K_{si}$ may result in large joint friction. As shown in Fig. \ref{fig3.3}, after $K_{si}$ has been loaded, applying a virtual external torque ${\tau}_{ei}={F}_{ei} R$ ($F_{ei}$ denotes the virtual external force, $R$ denotes the joint moment arm) to the joint will result in a passive rotation angle $\delta$ (rotates to the dashed position). $K_{si}$ refers to the level of effort required to produce a passive rotational angle $\delta$ with ${\tau}_{ei}$, which is:
\begin{equation}
K_{si}=\frac{{\tau}_{ei}}{\delta}=\frac{{F}_{ei}R}{\delta}\label{eq3.5}
\end{equation}

As described in Fig. \ref{fig3.2}(b) and Fig. \ref{fig3.3}, by applying ${\tau}_{ei}$, ${F}_{t1}$ will increase to ${F}'_{t1}$ while $d_{ji}$ on actuator 1, $d_{1i}$ to $d_{1i}=d_{si}+\delta R$. ${F}_{t2}$ will decrease to ${F}'_{t2}$ while $d_{2i}$ to $d_{2i}=d_{si}-\delta R$. ${F}_{ei}$ can be calculated as:
\begin{equation}
{F}_{ei}={F}'_{t1}-{F}'_{t2}+{F}_{f}\label{eq3.6}
\end{equation}

Where ${F}_{f}=\mu_{s} {F}_{tj}$ refers to the static frictional force acting on the joint, $\mu_s$ represents the static friction coefficient.

Combined \eqref{eq3.3} and \eqref{eq3.6}, ${F}_{ei}$ can be represented as:
\begin{equation}
{F}_{ei}=f_{di} (d_{si}+\delta R)-f_{di} (d_{si}-\delta R)+\mu_{s} {F}_{tj}\label{eq3.7}
\end{equation}

For different $d_{si}$, there are five stages when the joint is passively rotated.

The first stage is when $0 \leq d_{si} \leq \delta R$, ${\tau}_{ei}$ causes the elastic element in actuator 2 to resume its initial position. i.e. ${F}'_{t2}=0$. According to \eqref{eq3.7}, ${F}_{ei}$ at this stage is:
\begin{equation}
\left[
\begin{array}{c}
   {F}_{e1} \\
   {F}_{e2} \\
\end{array}
\right ]
=
\left[
\begin{array}{cc}
   k_{et1} & 1+\mu_{s}\\
   k_{et2} & 1+\mu_{s} \\
\end{array}
\right ]
\left[
\begin{array}{c}
\delta R \\
{F}_{tj}
\end{array}
\right ]
\label{eq3.8}
\end{equation}

Where $k_{et1}=k_{ts} k_{t1}/[k_{t1} (1-\mu_{p})+k_{ts}]$ and $k_{et2}=k_{cs} k_{t2}/(k_{t2}+k_{cs})$ are the stiffness of ICA and ECA during the springs working stage when the deformation of the tendon is considered.

The second stage is when $\delta R$\textless$d_{si} \leq d_{mi}-\delta R$, ${\tau}_{ei}$ causes the elastic element in actuator-1 further to deform but yet reach the maximum deformation. In this stage, the elastic element in actuator 2 resumes partial deformation and remains in a partially deformed state, unable to return to its original state. ${F}_{ei}$ at this stage is:
\begin{equation}
\left[
\begin{array}{c}
   {F}_{e1} \\
   {F}_{e2} \\
\end{array}
\right ]
=
\left[
\begin{array}{cc}
   k_{et1} & \mu_{s}\\
   k_{et2} & \mu_{s} \\
\end{array}
\right ]
\left[
\begin{array}{c}
2 \delta R \\
{F}_{tj}
\end{array}
\right ]
\label{eq3.9}
\end{equation}


{In the initial two phases, joint stiffness emanates from the deformation of the elastic component. The force-displacement relationship during these stages can be discerned by integrating the relevant parameters into the designated equation.} Practically, {complete relaxation of the tendon should be avoided to prevent the actuator from dislocation, this also impacts the control precision of the joints.} The second stage is the stage in which the joint stiffness can be controlled. The minimum and maximum controllable joint stiffnesses are denoted as $K_{smini}$, $K_{smaxi}$ with a range denoted as $\Delta K_{si}$. In the following three stages, the deformation of the tendon contributes to the joint stiffness adjustment. The tendon stiffness $k_{ti}$ is assumed to be constant, whereas practically $k_{ti}$ may vary. The following stages are therefore non-controllable joint stiffness stages.

The third stage is when $d_{mi}-\delta R$\textless$d_{si} \leq d_{mi}$, the elastic element in actuator 1 is fully deformed to the limited position, the tendon is stretched further. The elastic element of the actuator 2 resumes partial deformation. ${F}_{ei}$ at this stage is:
\begin{equation}
\small
\left[
\begin{array}{c}
   {F}_{e1} \\
   {F}_{e2} \\
\end{array}
\right ]
=
\left[
\begin{array}{ccc}
   {F}_{tm1} & \mu_{s}-1 & k_{1}\\
   {F}_{tm2} & \mu_{s}-1 & k_{2}\\
\end{array}
\right ]
\left[
\begin{array}{c}
1 \\ {F}_{tj} \\ \delta R
\end{array}
\right ]
+
\left[
\begin{array}{c}
F_{k13} \\
F_{k23} \\
\end{array}
\right ]
\label{eq3.10}
\end{equation}

Where $k_{1}=k_{et1}+k_{t1}$, $k_{2}=k_{et2}+k_{t2}$, $F_{k13}=k_{t1} (d_{s1}-d_{max1})$, $F_{k23}=k_{t2} (d_{s2}-d_{max2})$. 

The forth stage is when $d_{mi}$\textless$d_{si} \leq d_{mi}+\delta r$, the elastic element is in the limited position before $\tau_{ei}$ applied. ${\tau}_{ei}$ further stretch the tendon in actuator 1. The tendon in actuator 2 recovery from stretching and the elastic element resumes partial deformation. ${F}_{ei}$ at this stage is:
\begin{equation}
\small
\left[
\begin{array}{c}
   {F}_{e1} \\
   {F}_{e2} \\
\end{array}
\right ]
=
\left[
\begin{array}{ccc}
   -{F}_{tm1} & 1+\mu_{s} & k_{t1}\\
   -{F}_{tm2} & 1+\mu_{s} & k_{t2}\\
\end{array}
\right ]
\left[
\begin{array}{c}
1 \\ {F}_{tj} \\ \delta R
\end{array}
\right ]
+
\left[
\begin{array}{c}
F_{k14} \\
F_{k24} \\
\end{array}
\right ]
\label{eq3.11}
\end{equation}

Where $F_{k14}=(\delta R-d_{s1}+d_{max1}) k_{et1} \delta R$, $F_{k24}=(\delta R-d_{s2}+d_{max2}) k_{et2} \delta R$. 

For the last stage, when $d_{si}$ \textgreater $d_{mi}+\delta r$, the elastic element is in the limited position before ${\tau}_{ei}$ is applied. ${\tau}_{ei}$ further stretch the tendon in actuator 1. In actuator 2 the tendon recovers partial deformation. ${F}_{ei}$ at this stage is:
\begin{equation}
{F}_{ei}=2 \delta R k_{ti}+\mu_{s} {F}_{tj}\label{eq3.12}
\end{equation}

Combining five stages, ${F}_{ei}$ can be calculated as a function:
\begin{equation}
{F}_{ei}=f_{ei} (\delta,d_{si})\label{eq3.13}
\end{equation}

Combining \eqref{eq3.3}, \eqref{eq3.5} and \eqref{eq3.13}, $K_{si}$ loaded to the joints at different $d_{si}$ can be expressed as a function:
\begin{equation}
K_{si}=f_{Ki} (\delta,d_{si})\label{eq3.14}
\end{equation}

\subsection{Maximum allowable joint acceleration}

After joint stiffness $K_{si}$ is determined, the actuators will rotate the joint as the second step in Fig. \ref{fig3.2}(c). Since the actuators are not fixed, it is critical to prevent the actuators from loosening or even dislocating during joint motion. This subsection calculates the relationship between the maximum allowable angular acceleration $\ddot \theta_{max}$ (to avoid tendon slackening) and $d_{si}$ during the joint rotation.

Motor 1 continues to rotate at an angle $\alpha_{ri}$ while motor 2 continues to rotate $-\alpha_{ri}$, where $\alpha_{ri}$ is the driving angle. Assuming the effects of gravity and friction are neglected, the test joint would be actuated to the desired position $\theta$ (represented by solid position in Fig. \ref{fig3.2}(c)) if $\alpha_{ri} r=\theta R$.

During the acceleration of the joint rotation, {the joint will be passively rotated by angle $\delta_{r}$} due to the mass moment of inertia. The elastic element in actuator 1 will be further deformed and the tendon displacement will increase to $d_{1i}=d_{si}+\delta_{r} R$. The elastic element in actuator 2 will partially recover its deformation and $d_{2i}$ will decrease to $d_{2i}=d_{si}-\delta_{r} R$. If the joint angular acceleration $\ddot \theta$ exceeds the allowable maximum value, the elastic element of actuator 2 will restore to its initial position and the tendon will no longer be tensioned (${F}_{t2}=0$). At this point the tendon displacement on actuator 2, $d_{2i}$ will decrease from $d_{si}$ to $0$, therefore:
\begin{equation}
d_{si}-\delta_{r} R=0\label{eq3.15}
\end{equation}

According to \eqref{eq3.13}, the virtual external torque required to passively rotate the joint at angle $\delta_{r}$ is:
\begin{equation}
{\tau}_{ei}={F}_{ei} R=f_{ei} (\delta_{r},d_{si} )R\label{eq3.16}
\end{equation}

In order to avoid the tendon slack on actuator 2, ${\tau}_{ei}$ is related to the maximum allowable angular acceleration $\ddot \theta_{max}$ as:
\begin{equation}
{\tau}_{ei}=I \ddot \theta_{max} \label{eq3.17}
\end{equation}

Where $I$ is the rotational inertia of the joint.

$\ddot \theta_{max}$ is related to $d_{si}$. Combining \eqref{eq3.15}, \eqref{eq3.16} and \eqref{eq3.17}, the relationship can be obtained as:
\begin{equation}
\ddot \theta_{max}=\frac{f_{ei} (d_{si}/R,d_{si})R}{I} \label{eq3.18}
\end{equation}

Here we only consider the situation where the actuator is in the stage when the elastic element operates, so that $d_{si} \leq d_{mi}$.

\subsection{Joint torque control}

Figure \ref{fig3.2}(d) illustrates a joint that is restricted and $K_{si}$ is maintained. In this setup, while motor 1 keeps rotating by an angle $\alpha_{r}$, motor 2 remains stationary. This causes $d_{1i}$ to increase by $d_{ti}$, such that $d_{1i}=d_{si}+d_{ti}$, where $d_{ti}$ represents the torque displacement, which is equal to $\alpha_{r} r$. The force on actuator 2 remains constant at ${F}_{t2}$, while ${F}_{t1}$ increases to ${F}''_{t1}$. Using equation \eqref{eq3.3}, we can deduce the tension difference between the two actuators $F_{ti}$ as:
\begin{equation}
{F}_{ti}=f_{di} (d_{si}+d_{ti})-f_{di} (d_{si}-d_{ti})-{F}_{f}\label{eq3.19}
\end{equation}

Thus, the joint should output torque ${\tau}_{ti}={F}_{ti} R$, which is a function of $d_{ti}$ and $d_{si}$:
\begin{equation}
{\tau}_{ti}=f_{tdi} (d_{si},d_{ti})\label{eq3.22}
\end{equation}

When $d_{si}+d_{ti} \leq d_{mi}$, where the elastic elements are in the working stage, it is considered as the torque controllable stage. {During this stage, the maximum joint torque ${\tau}_{tmi}$ is related to $d_{si}$:}
\begin{equation}
{\tau}_{tmi}=f_{tdi} (d_{si},d_{mi}-d_{si})\label{eq3.23}
\end{equation}

The above is to output ${\tau}_{ti}$ while maintaining $K_{si}$, i.e. actuator 2 remains stationary and actuator 1 is in operation. It is also possible to produce a larger ${\tau}_{ti}$ by increasing $d_{1i}$ and decreasing $d_{2i}$ simultaneously when the joint is restricted. At this point, ${\tau}_{tmi}$ will be constant, independent from $d_{si}$ (friction is not considered). If only consider the stages when elastic elements are involved, ${\tau}_{tmi}$ occurs when ${F}_{t1}={F}_{tmi}$ and ${F}_{t2}=0$ as:
\begin{equation}
{\tau}_{tmi}=R{F}_{tmi} \label{eq3.24}
\end{equation}

\section{Prototypes and comparisons}

\begin{figure}[htbp]
\centerline{\includegraphics[width=0.6\textwidth]{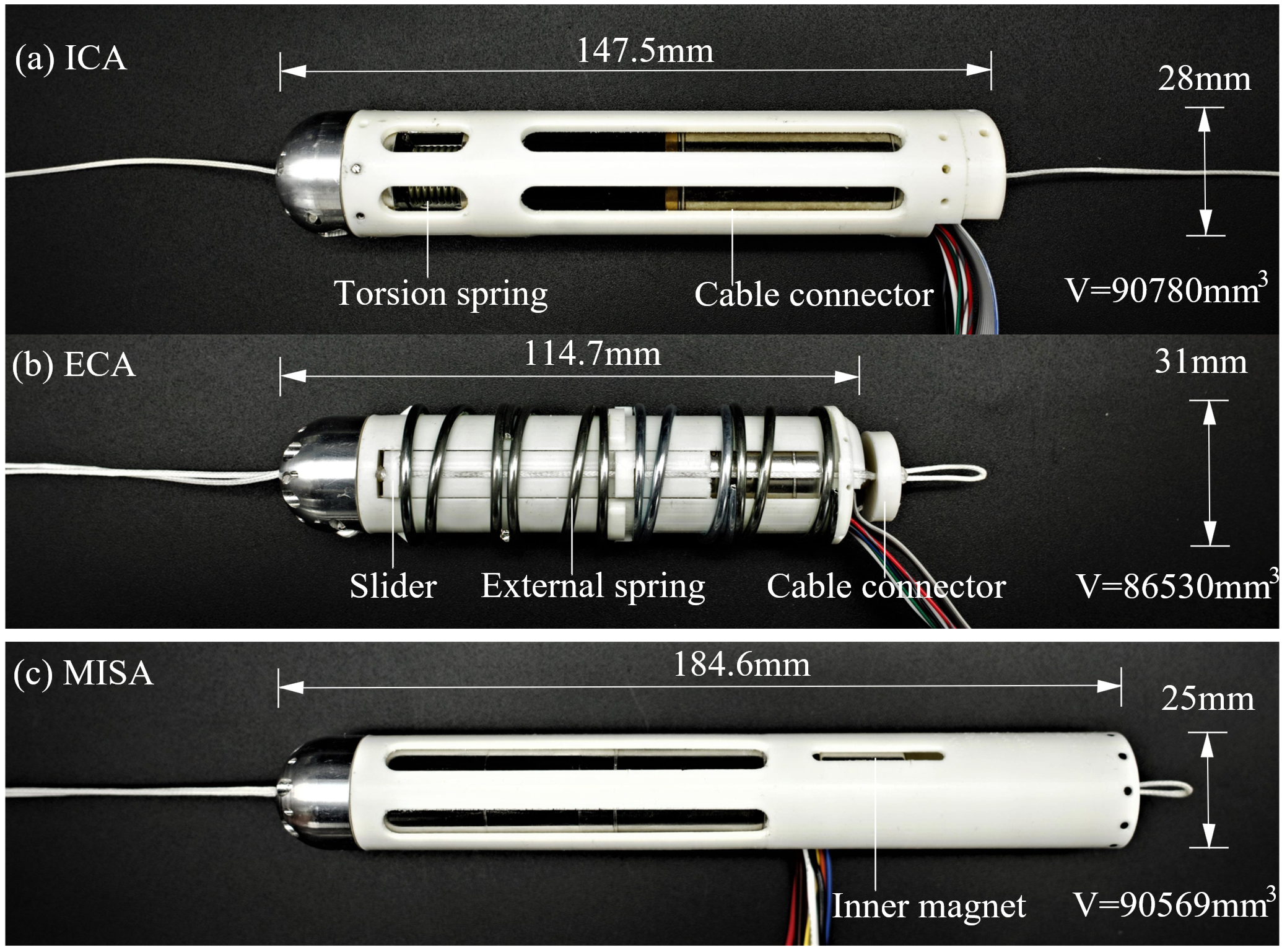}}
\caption{The prototypes of the three compliant actuators: (a) ICA; (b) ECA; (c) MISA\cite{yang2023novel}.}
\label{fig3.4}
\end{figure}

To enable a clear comparison, the ICA, ECA and the previously introduced MISA\cite{yang2023novel} were developed utilizing identical motors: {Maxon brushless motors ECX TORQUE 22 M (24V, 30/43W) with GPX 22 mm 62:1 planetary gearbox}. An output pulley with a radius of $r=5$mm is incorporated in each design. The prototypes are depicted in Figure \ref{fig3.4}, while their key parameters are tabulated in Table \ref{tab3.1}. Despite varying in diameter and length, the volumes of the three actuators are relatively similar.

\begin{table}[htbp]
\caption{Performance of the prototypes}
\footnotesize
\begin{center}
\begin{tabular}{l c c c}
\hline
 & ECA & ICA & MISA\\
\hline
Rated force (N) & 250 & 125 & 250   \\

Rated speed (mm/s) & 110 & 220 & 110  \\

Diameter (mm) & 31 & 28 & 25  \\

Length (mm) & 114.7 & 147.5 & 184.6  \\

Mass (g)  & 295 & 280 & 335  \\

Volume (mm$^{3}$) & 86530 & 90780 & 90569  \\
\hline
\end{tabular}
\label{tab3.1}
\end{center}
\end{table}

\begin{figure*}[htbp]
\centerline{\includegraphics[width=0.7\textwidth]{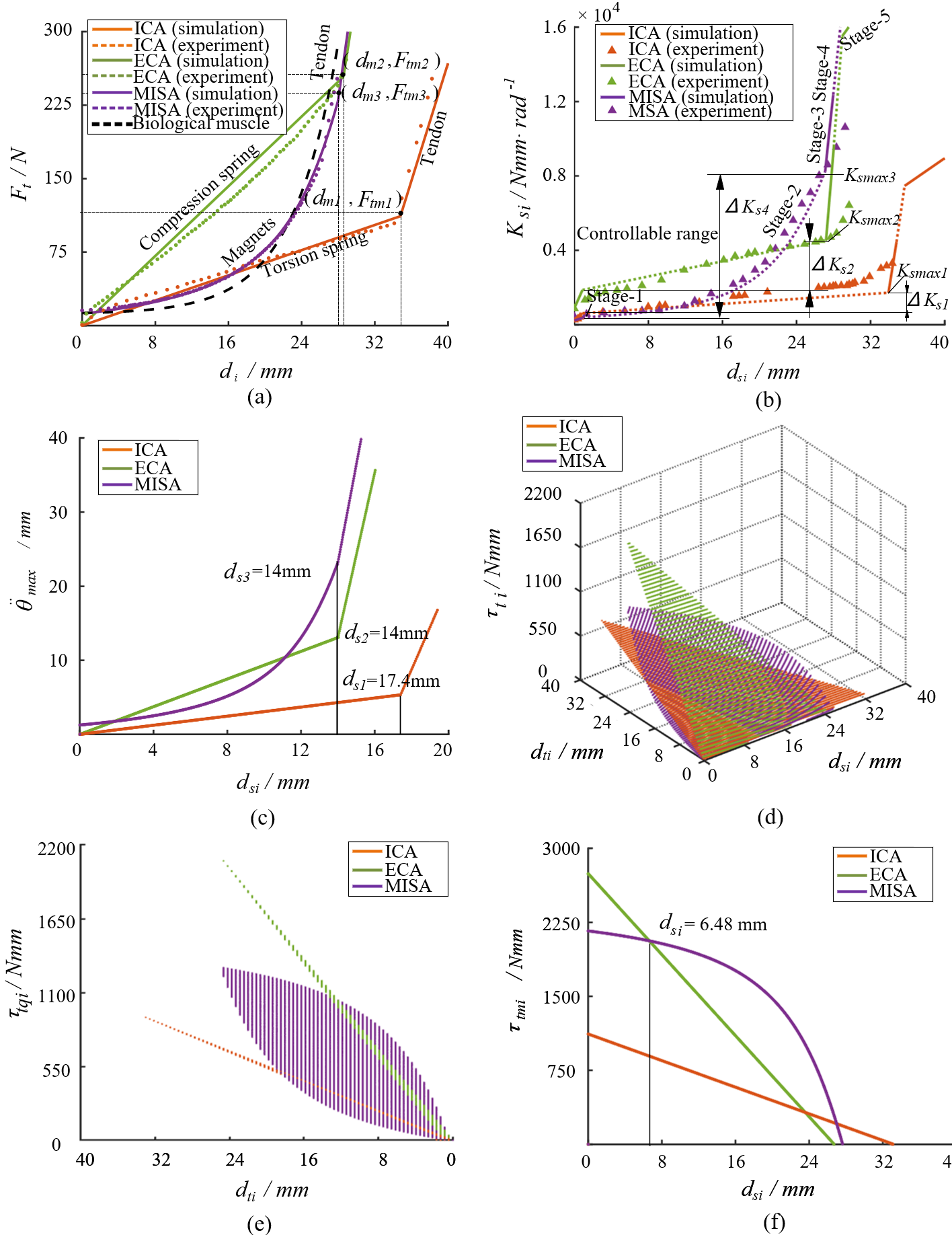}}
\caption{(a) The simulation and experimental results of the three compliant actuators, ${F}_{t}=f_{di} (d_{i})$, and the tension-length curve of the biological muscle; (b) $K_{si}=f_{Ki} (\delta=0.087,d_{si})$; (c) Simulation result of the relation between $\ddot \theta_{max}$ and $d_{si}$; (d) ${\tau}_{ti}=f_{tdi} (d_{si},d_{ti})$ ($R$ = 10 mm); (e) side view of ${\tau}_{ti}=f_{tdi} (d_{si},d_{ti})$ ($R$ = 10 mm); (f) relationship between ${\tau}_{tmi}$ and $d_{si}$. }
\label{fig3.5}
\end{figure*}

Both actuators use the same tendon, which is a braided fishing line with an outer diameter of 1.0 mm and a maximum force of 90.72 kg. The tendon's spring constant $k_t$ was determined to be 30 N/mm through experimentation, and its length is fixed at 200 mm. However, the number of tendons used in each actuator differs, resulting in different values of $k_{ti}$.

Based on Table \ref{tab3.1}, ICA is designed to withstand a rated force of 125 N and an ultimate motor torque of 1.25 Nm. However, as a result of the limitations imposed by size, a torsion spring (elasticity coefficient: $k_{e1} = 508$ Nmm/rad, maximum deformation: 1 rad) was selected. This corresponds to an equivalent elasticity coefficient of $k_{ts} = k_{e1}/2\pi r^{2} = 3.236$ N/mm due to the tendon displacement. The torsion spring reaches the maximum deformation when ${F}_{tm1} = 112.4$ N. Since minimizing the actuator diameter require narrow access inside the shell, the tendons use a single strand of fishing line with $k_{t1} = 30$ N/mm and $d_{m1} = 34.8$ mm.

ECA is designed to withstand a rated force of 250 N. A compression spring with an elasticity coefficient of $k_{e2} = k_{cs} = 10.44$ mm/N and a maximum compression of 24.34 mm is selected. The spring is compressed to the solid position when ${F}_{tm2} = 254$ N, providing protection even if the motor reaches the rated torque. The tendon in ECA consists of two strands, and the elasticity coefficient is $k_{t2} = 60$ N/mm, resulting in $d_{m2} = 28.5$ mm.

\subsection{Comparison of simulation results}

\begin{figure*}[htb]
\centerline{\includegraphics[width=1\textwidth]{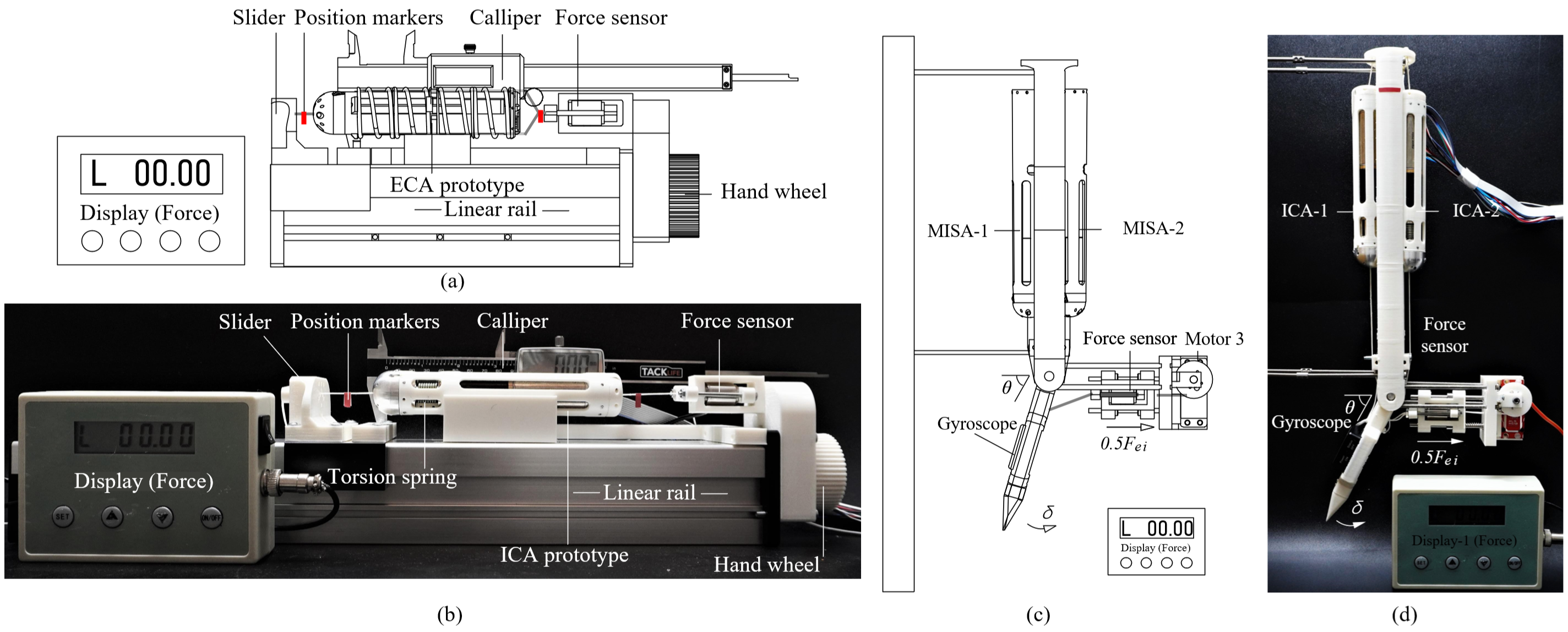}}
\caption{(a) The schematic diagram of the force-displacement relationship test rig. (b) Test rig setup for force-displacement relationship test; (c) The schematic diagram of the variable joint stiffness test rig. (d) Test rig set-up for the adjustable joint stiffness test.}
\label{fig3.8}
\end{figure*}

\begin{table}[htbp]
\caption{Parameters of the prototypes}\label{tab3.2}
\footnotesize
\centering
\begin{threeparttable}
\begin{tabular}{l l l }
\toprule
 & ECA & ICA \\
\midrule
$k_{ei}$ (N/mm)& $k_{cs}$ = 10.4 & $k_{ts}$ = 3.2  \\

${F}_{tmi}$ (N)& 252.9 & 112.4   \\

$d_{mi}$ (mm)& 28.5 & 34.8   \\

$k_{ti}$ (N/mm)& 60 & 30   \\

$K_{smini}$$^1$ (Nm/rad)& 1.82 & 0.60  \\

$K_{smaxi}$$^1$ (Nm/rad)& 3.12 & 1.15  \\

$\Delta K_{si}$$^1$ (Nm/rad)& 1.30 & 0.52  \\

$\tau_{tmi}$$^2$ (Nm) & 2.77  & 1.12  \\

\bottomrule
\end{tabular}
\begin{tablenotes}
\item[1]$\delta$ = 0.087 rad, $\mu_e$ = 0.1, $\mu_s$ = 0.1, $r$ = 5 mm, $R$ = 10 mm.
\item[2]${\tau}_{tmi}$ is calculated while Motor-2 remains stationary.
\end{tablenotes}
\end{threeparttable}
\end{table}

In this subsection, the simulation results of the ICA, ECA and MISA\cite{yang2023novel} will be presented and compared. ICA is represented in orange, ECA in green and MISA in purple.

The parameters of the actuators are presented in Table \ref{tab3.2}. Fig. \ref{fig3.5}(a) displays the simulated force-displacement relation, ${F}_{t}=f_{di} (d_{i})$. The curve for the biological muscle is represented by the dashed blue line in Fig. \ref{fig3.5}(a), indicating a rise in stiffness as displacement increases. For acquisition details, refer to our previous work\cite{yang2023novel}. This non-linear behaviour can provide quick protection against unexpected impacts. MISA aims to replicate the non-linear stiffness exhibited by biological muscles. By comparing the solid and dashed blue curve in Fig. \ref{fig3.5}(a), it is demonstrated that the trends in MISA closely coincide with that of biological muscle. In contrast, the force-displacement relations for the ICA and ECA are linear.

A pair of ICAs, ECAs and MISAs were applied to the same test joint ($R$ = 10 mm) respectively. By adjusting $d_{si}$, according to \eqref{eq3.14}, different joint stiffness $K_{si}$ can be provided to the joint. $K_{si}$ is related to $\delta$ and $d_{si}$. In Fig. \ref{fig3.5}(b), the relationship between $K_{si}$ and $d_{si}$ when $\delta$ = 0.087 rad is illustrated, i.e. $K_{si}=f_{Ki}$ ($\delta$ = 0.087, $d_{si}$). As described in the previous section, five stages of $K_{si}$ are distinguished by different line types. The second stage, which is the controllable joint stiffness stage, is indicated by the dotted lines in Fig. \ref{fig3.5}(c).


According to Fig. \ref{fig3.5}(a), ECA and MISA share similar ${F}_{tmi}({F}_{tm2}\approx {F}_{tm3})$. But using the same calculation methodology, it can be observed from Fig. \ref{fig3.5}(b) that the joint using MISAs has the ability to realize a board range of joint stiffness compared to ECAs, $\Delta K_{s3}$ = 7850 Nmm/rad, 2.9 times the $\Delta K_{s2}$ = 2640 Nmm/rad of ECAs. The maximum controllable joint stiffness achievable with the MISA is $K_{smax3}$ = 8235 Nmm/rad, 1.83 times the $K_{smax2}$ = 4507 Nmm/rad of the ECA. The minimum controllable joint stiffness for the joint with ECAs is 1867 Nmm/rad, lower joint stiffnesses cannot be achieved. 

Fig. \ref{fig3.5}(c) shows the maximum allowable angular acceleration $\ddot \theta_{max}$ during joint rotation with different $d_{si}$ (different $K_{si}$). $\ddot \theta_{max}$ is positively {related} to $d_{si}$. A marked escalation is observed in $\ddot \theta_{max}$ once $d_{si}$ surpasses a specified threshold.
($d_{s1}$ \textgreater 14 mm, $d_{s2}$ \textgreater 17.4 mm, $d_{s3}$ \textgreater 14 mm). {This occurs as the tendon in actuator-1 experiences stretching before the tendon in actuator-2 has fully slackened, a situation that is further compounded by the tendon's high elasticity coefficient.}

${\tau}_{ti}$, in relation to $d_{si}$ and $d_{ti}$, ${\tau}_{ti}=f_{tdi}$ ($d_{si}$, $d_{ti}$) is shown in Fig. \ref{fig3.5}(d) and (e). Using this relationship, joint torque control can be achieved. If only consider the elastic element working stage, i.e. $d_{si}+d_{ti} \leq d_{mi}$, the maximum output joint torque ${\tau}_{tmi}$  is only related to $d_{si}$ as shown in Fig. \ref{fig3.5}(f). It can be seen that ${\tau}_{tmi}$ decreases if $d_{si}$ increases. When $d_{si}$ \textless 6.48 mm, the joint with ECAs is able to output a larger torque than that with MISAs, ${\tau}_{tm2} < {\tau}_{tm3}$. 




\subsection{Verification of Force-Displacement correlation}

The test-rig setup is depicted in Fig. \ref{fig3.8} (a) and (b), in which the actuator tendon was secured at one end to a force sensor (Dysensor, DYZ-102, 0-500 N), and at the other end was connected to the slider. As the slider was free to move along the linear rail, the position was adjustable using a hand wheel. Position markers were employed to monitor the tendon displacement. The motor was energized during the experiment but remained stationary. The experimental data were recorded by the force sensor and the calliper at various positions of the slider. {The associated test videos are provided in the supplemental materials.} The obtained results for the actuators are displayed in Fig. \ref{fig3.5}(a), which are consistent with the simulation results.

\subsection{Verification of joint stiffness-displacement relationship}

Figs. \ref{fig3.8}(c) and (d) shows the test set-up for evaluating joint stiffness using a pair of antagonistic actuators applied to a test joint. Motor 3 is responsible for moving the force sensor along a linear rail, which is attached to the driven arm via cables. The movement of the force sensor can rotate the driven arm counterclockwise. The force value is displayed on the display, and {a commercialized gyroscope sensor (Brand: Wit-motion, Model: WT901BLECL, Chip: MPU9250)} is attached to record the the driven arm's position. The joint stiffness-displacement relation can be tested in the following steps:

Step 1: The joint is driven to the target angle (70$^{\circ}$) by the two ICAs without any deformation of the elastic elements.

Step 2: Control Motor 3 rotates and shifts the force sensor gradually towards the right. The external force from the force sensor causes the driven arm to rotate counterclockwise. The force is recorded when the driven arm reaches the desired position (75$^{\circ}$). Then control Motor 3 to return to its initial position.

Step 3: Both actuators 1 and 2 are controlled to generate different tendon displacements $d_{si}$ while maintaining the joint position constant at the value achieved in the first step. Then the second step is repeated, and the force readings are recorded throughout the test.

The test results provide the relationship between ${F}_{ei}$ and $d_{si}$ for each actuator, which in turn determines the relationship between $K_{si}$ and $d_{si}$. {The associated test videos are provided in the supplemental materials.} The plot of these results for the three actuators is presented in Fig. \ref{fig3.5}(b).

\begin{figure*}[htbp]
\centerline{\includegraphics[width=1\textwidth]{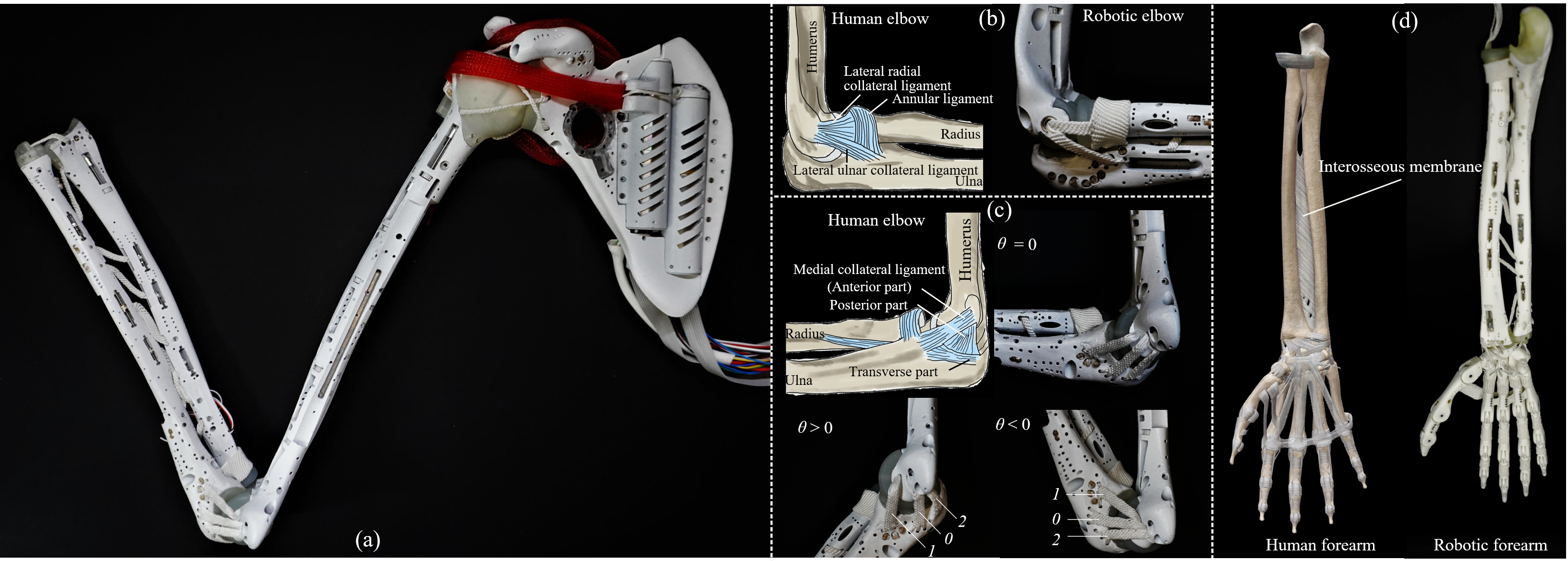}}
\caption{{(a) Robotic arm prototype based on human skeleton system. The joint is articulated by ligaments. (b) The lateral collateral ligament and annular ligament of the human elbow and the robotic elbow; (c) Strain varies on each part of the medial collateral ligament when the joint angle changes. Part 1 is under tension when $\theta>0$, part 2 is under tension when $\theta<0$; (d) A robotic hand is attached to the arm, a skeletal ligament system similar to that of the human forearm is replicated in the robotic arm.}}
\label{fig3.17}
\end{figure*}

\section{Application of compliant actuators on the biomimetic robotic arm}

\begin{figure*}[htbp]
\centerline{\includegraphics[width=0.8\textwidth]{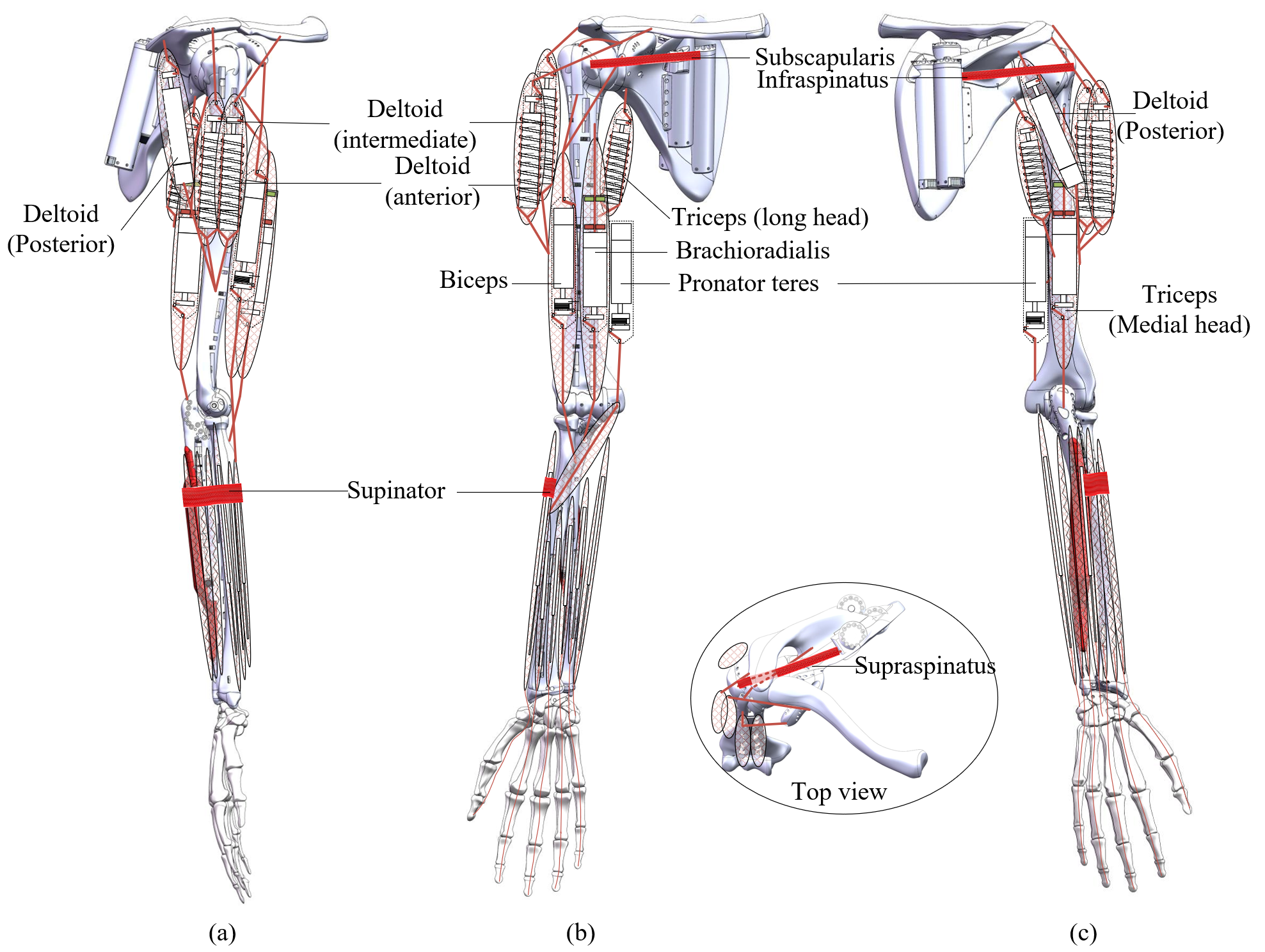}}
\caption{{Tendon and muscle arrangement of the proposed robotic arm (right). (a) Right view; (b) Front view; (c) Rear view.}}
\label{fig3.16}
\end{figure*}

The human body and traditional robots differ fundamentally in the type and function of their respective tissues. Traditional robots consist mainly of rigid materials, such as metal and plastic, while the human body is made up of soft tissues, including muscles, tendons, fat, cartilage, and ligaments. This key difference means that the human body possesses unique characteristics such as both tensile and compressive properties, including variable joint stiffness, flexibility, damping, compliance, and more, which are critical for exceptional performance in tasks such as fine motor skills, dexterity, and adaptability to a wide range of environments.

The human musculoskeletal system serves as an ideal model for designing a robotic arm. Fig. \ref{fig3.17}(a) shows a robotic arm that imitates the skeletal-ligament system of the biological arm, replicating the bones, ligaments, and cartilage of the human arm. For example, the elbow joint is stabilized by ligaments (medial collateral ligament and lateral collateral ligament) instead of being articulated with a shaft. To replicate the medial collateral ligament in the robotic arm prototype, it is divided into three parts: the anterior part (1), middle part (2), and posterior part (3), as shown in Fig. \ref{fig3.17}(b). At the initial position ($\theta$ = 0), all three parts of the ligament are at their original length. This arrangement allows the middle part (2) to provide stability throughout the rotation of the elbow, while the tension on the posterior part (3) increases sharply as the joint approaches full flexion ($\theta$ \textless 0), preventing joint dislocation. Similarly, as the joint approaches full extension ($\theta$ \textgreater 0), the tension on the anterior part (1) increases dramatically to limit the range of motion. The annular ligament and lateral collateral ligament are also replicated in the elbow as shown in Fig. \ref{fig3.17}(c). These ligaments are knitted with high-strength fibre and exhibit spring-like properties, providing biomimetic features to the robotic arm\cite{lu2022reproduction}. To reduce friction and provide a joint interface that mimics biology, the elbow contact surface has been coated with artificial cartilage made of durable V2 resin {(made by Formlabs, Elongation at break: 55\%, Ultimate Tensile strength: 28 MPa, Tensile modulus: 1 GPa)}. Due to the soft tissues in the elbow, the stiffness and damping of the joint vary during the joint angle change.

The forearm of the human arm is composed of the radius and ulna, which is also the case for the proposed robotic arm. Pronation/supination is achieved by the radius rotating around the ulna. The radius and ulna are articulated by the interosseous membrane, which is also duplicated in the proposed robotic arm, as shown in Fig. \ref{fig3.17}(d).

The actuators proposed in this paper have been developed primarily for tendon-driven robotic arms featuring highly biomimetic. The robotic arm presented serves as the target robot, which not only demands an appearance that resembles the human body but also a design that adheres to the underlying principles of the human body. 

\begin{figure*}[htbp]
\centerline{\includegraphics[width=0.7\textwidth]{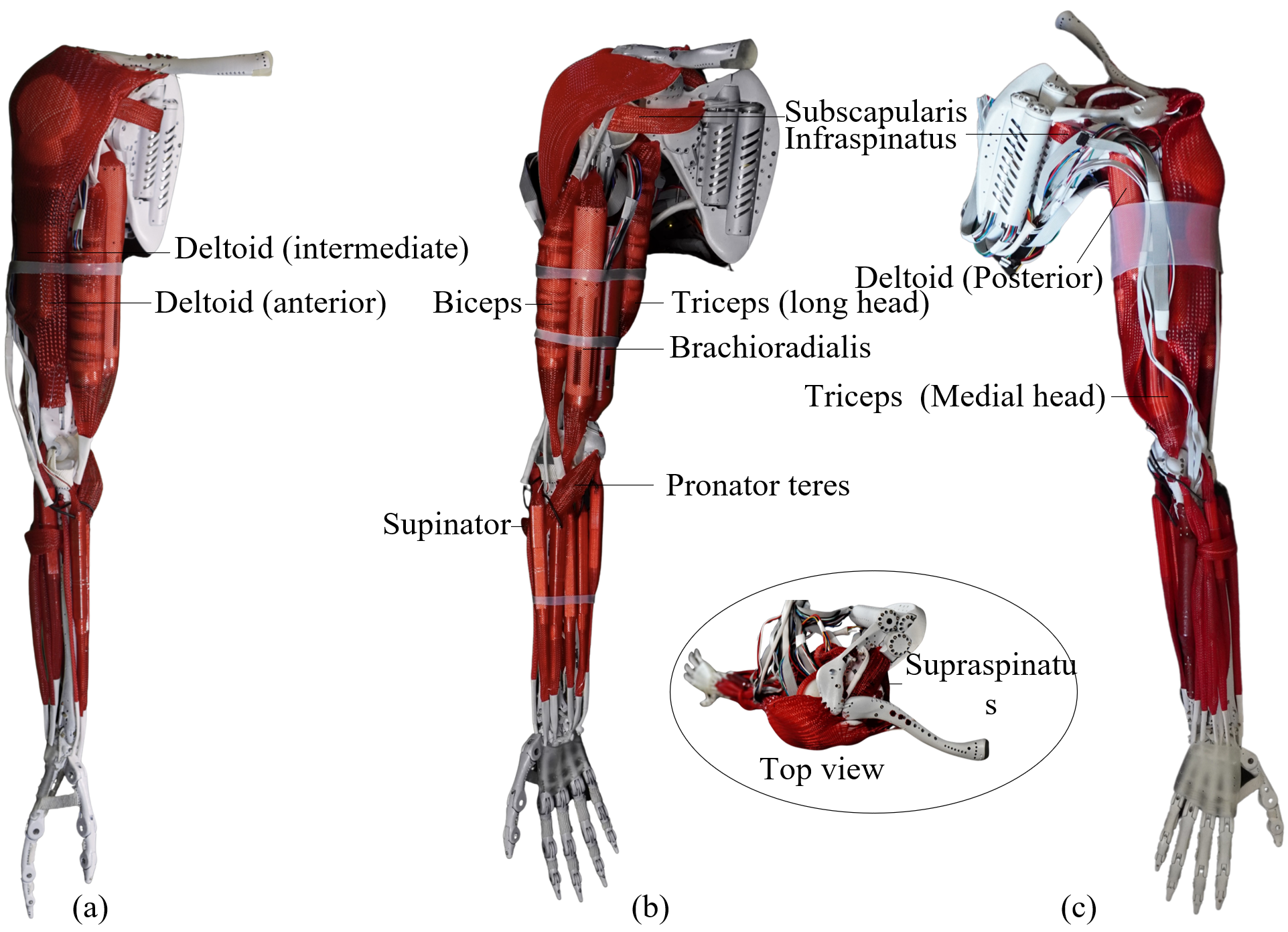}}
\caption{{The proposed robotic arm. (a) Right view; (b) Front view; (c) Rear view.}}
\label{fig3.161}
\end{figure*}

The robotic arm is developed by incorporating nine compliant actuators, as depicted in Fig. \ref{fig3.16}. Two MISAs work as the brachioradialis and medial head of the triceps. Two ICAs function as the pronator teres and supinator, while an ECA works as the biceps to assist in forearm rotation (the biceps could also assist with elbow actuation). Additionally, another ECA serves as the long head of the triceps for shoulder adduction and three ECAs work as the Deltoid. 12 major muscles are applied to the robotic arm, as listed in Table \ref{tab5.2}. The actuators responsible for driving the subscapularis, infraspinatus, supraspinatus, wrist flexor/extensor, and abductor/adductor, are not designed as compliant actuators. The robotic arm prototype is shown in Fig. \ref{fig3.161}.

\begin{table}[htb]
\caption{{Artificial muscles applied in the proposed robotic arm.}}
\footnotesize
\begin{center}
\begin{tabular}{l l l}
\toprule
Joint & Muscle & \makecell[c]{Type}\\
\midrule
Shoulder&Deltoid (anterior)     & ECA   \\
&Deltoid (intermediate) & ECA  \\
&Deltoid (posterior)    & ECA    \\
&Subscapularis          & Without compliance       \\
&Infraspinatus          & Without compliance        \\
&Supraspinatus          & Without compliance        \\
&Triceps (Long head)    & ECA     \\
\midrule
Elbow and forearm&Biceps                & ECA \\
&Brachioradialis       & MISA \\
&Triceps (Medial head) & MISA \\
&Pronator teres        & ICA \\
&Supinator             & ICA     \\
\midrule
Wrist&Wrist flexor/extensor   & Without compliance            \\
&Wrist abductor/adductor & Without compliance            \\
\bottomrule
\end{tabular}
\label{tab5.2}
\end{center}
\end{table}

\subsection{Exploring joint flexibility and workspace efficiency}

\begin{figure}[htb]
\centerline{\includegraphics[width=1\textwidth]{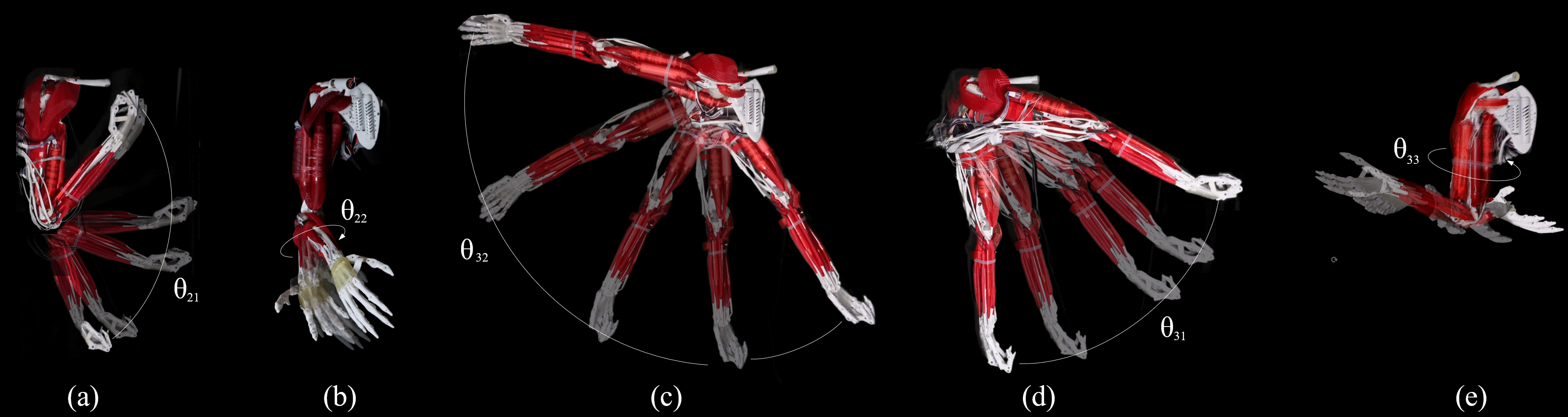}}
\caption{Range of motion test for each motion in the proposed robotic arm. (a) Elbow flexion/extension; (b) Forearm rotation; (c)Glenohumeral joint adduction/abduction; (d) Glenohumeral joint flexion; (e) Glenohumeral joint rotation.}
\label{fig8.3}
\end{figure}

\begin{figure}[htb]
\centerline{\includegraphics[width=0.65\textwidth]{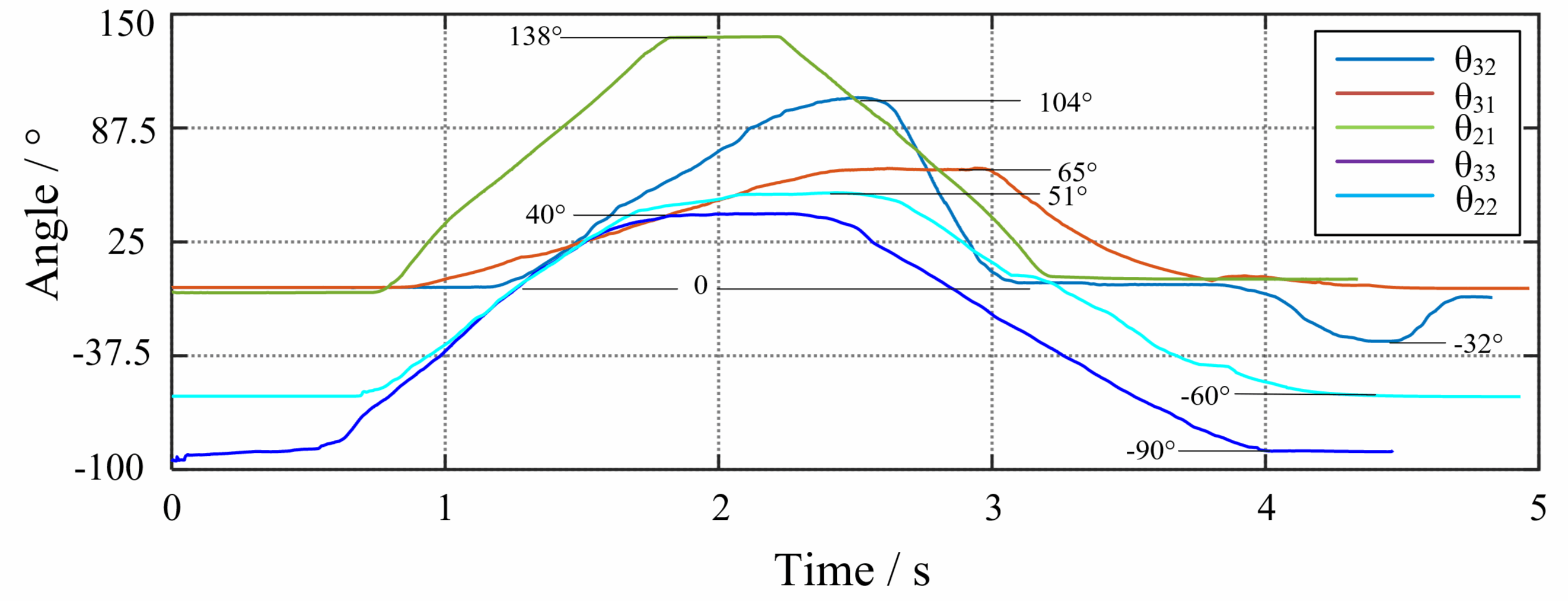}}
\caption{Data recorded of the range of motion test.}
\label{fig8.1}
\end{figure}

The elbow, forearm, wrist, and glenohumeral joint are designed with few joints to achieve multiple degrees of freedom and a large range of motion, reflecting the compactness of the human upper limb. Given the irregular geometrical nature of the joint surfaces, deriving the joint motion angles via simulation proves challenging. Consequently, an experimental approach was employed to ascertain the range of motion for each joint. To record each active joint motion, the scapula of the robotic arm is fixed to the platform, and a gyroscope is used. Before each experiment, the gyroscope is calibrated, and its position is modified so that each measured joint motion corresponds to a change in the x-axis rotation angle. The test results are shown in Fig. \ref{fig8.3} and Fig. \ref{fig8.1}, and the recorded ranges are compared with the data for the human arm, which is presented in Table. \ref{tab6.2}.

\begin{table}[htb]
\centering
\caption{The joint range of motion of the robotic arm and human arm (data from \cite{sheikhzadeh2008three}, \cite{asano2019musculoskeletal}).}
\footnotesize
\begin{center}
\begin{tabular}{l l l l}
\toprule
Motion group & Symbols & Joint range of motion (robotic arm) & human arm\\
\midrule
Glenohumeral extension (-) / flexion (+) & $\theta_{31}$ & -40*-65\degree & -60\degree-167\degree\\
Glenohumeral adduction (-) / abduction (+) & $\theta_{32}$ & -32\degree-104\degree & -29\degree-100\degree \\
Glenohumeral internal (-) / external (+) rotation & $\theta_{33}$ & -90\degree-40\degree & -97\degree-34\degree  \\
Elbow extension (-) / flexion (+) & $\theta_{21}$ & 0-138\degree & 0-146\degree \\
Forearm pronation (-) / supination (+) & $\theta_{22}$ & -60\degree-65\degree & -70\degree-70\degree\\
\bottomrule
\end{tabular}
\begin{tablenotes}
      \small
      \item[1]*The position of the glenohumeral joint extension was measured in the skeletal system.
\end{tablenotes}
\label{tab6.2}
\end{center}
\end{table}

\begin{figure}[htbp]
\centerline{\includegraphics[width=0.42\textwidth]{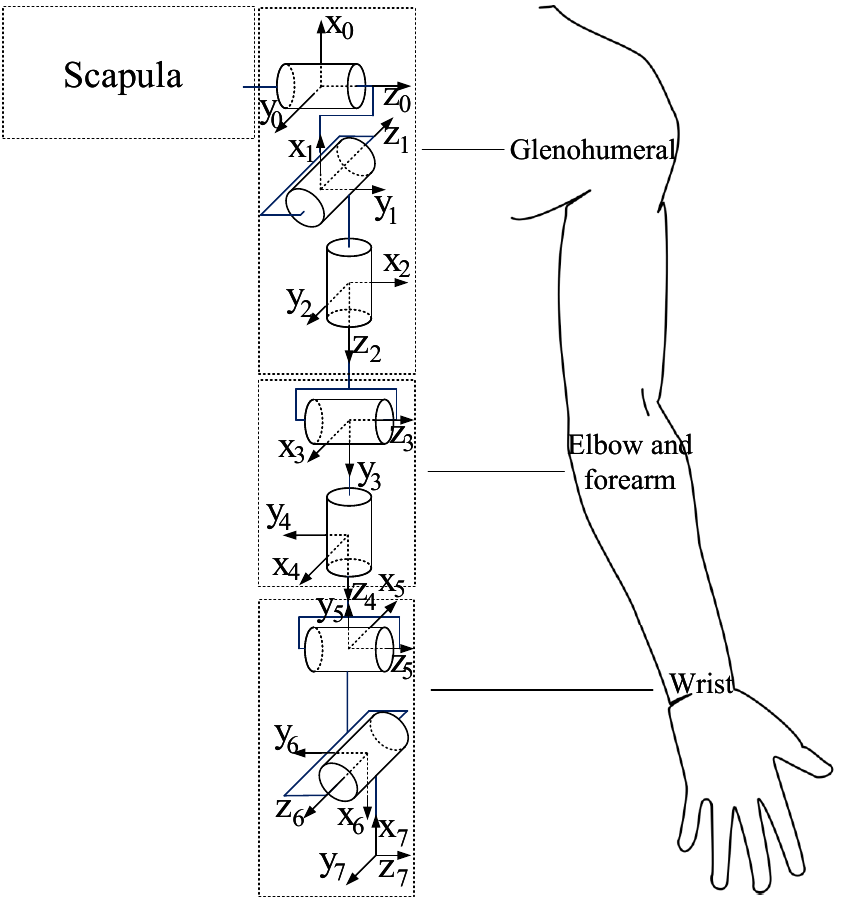}}
\caption{{The kinematic model of the proposed robotic arm (the hand is not included).}}
\label{fig3.18}
\end{figure}

The observed limitations within the prototype pertain to the omission of pectoralis major and latissimus dorsi, restricting the glenohumeral joint's flexion and extension range. Moreover, actuator positioning around the forearm for finger and wrist actuation slightly impedes forearm pronation. Significantly, the prototype's stationary scapula curtails shoulder complex abduction/adduction, mirroring a substantial reduction compared to human arm mobility. This fixed scapula position further hinders rearward extension of the glenohumeral joint, collectively delineating the prototype's biomechanical divergences and underscoring the necessity for refined actuator and musculoskeletal design to better emulate human motion intricacies.

The kinematic model of the robotic arm is presented in Fig. \ref{fig3.18}, which comprises 7 revolute joints. As the scapula motion in the shoulder complex was not considered in the early prototype, the first three joints account for the motion of the glenohumeral joint, which is well-modelled as a ball-and-socket joint. The fourth joint represents the elbow, while the fifth joint is for forearm rotation. The last two joints contribute to a 2-DOF wrist. Although there is a slight misalignment between the axis of forearm rotation and the wrist motion, this is disregarded in this model. According to the Denavit-Hartenberg convention, coordinate frames are attached to the joints as shown in Fig. \ref{fig3.18}, and the D-H parameters of the kinematic chain are presented in Table \ref{tab5.4}.

\begin{table}[htb]
\caption{Denavit-Hartenberg parameters of the kinematic model.}
\footnotesize
\begin{center}
\begin{tabular}{c c c c c}
\toprule
axis, i & $a_{i}$ & $d_{i}$ & $\alpha_{i}$ & $\theta_{i}$ \\
\hline



1 & 0 & 0 & $\pi/2$  & $\theta_{31}$ \\

2 & 0 & 0 & $-\pi/2$ & $\pi/2-\theta_{32}$ \\

3 & 0 &$b$& $\pi/2$  & $\pi/2+\theta_{33}$ \\

4 & 0 & 0 & $-\pi/2$ & $\theta_{21}$  \\
5 & 0 & c & $-\pi/2$ & $\pi+\theta_{22}$  \\
6 & 0 & 0 & $-\pi/2$ & $-\pi/2-\theta_{11}$  \\
7 & 0 & d & $\pi/2$  & $\theta_{12}$  \\
\bottomrule
\end{tabular}
\label{tab5.4}
\end{center}
\end{table}

\begin{figure}[htb]
\centerline{\includegraphics[width=1\textwidth]{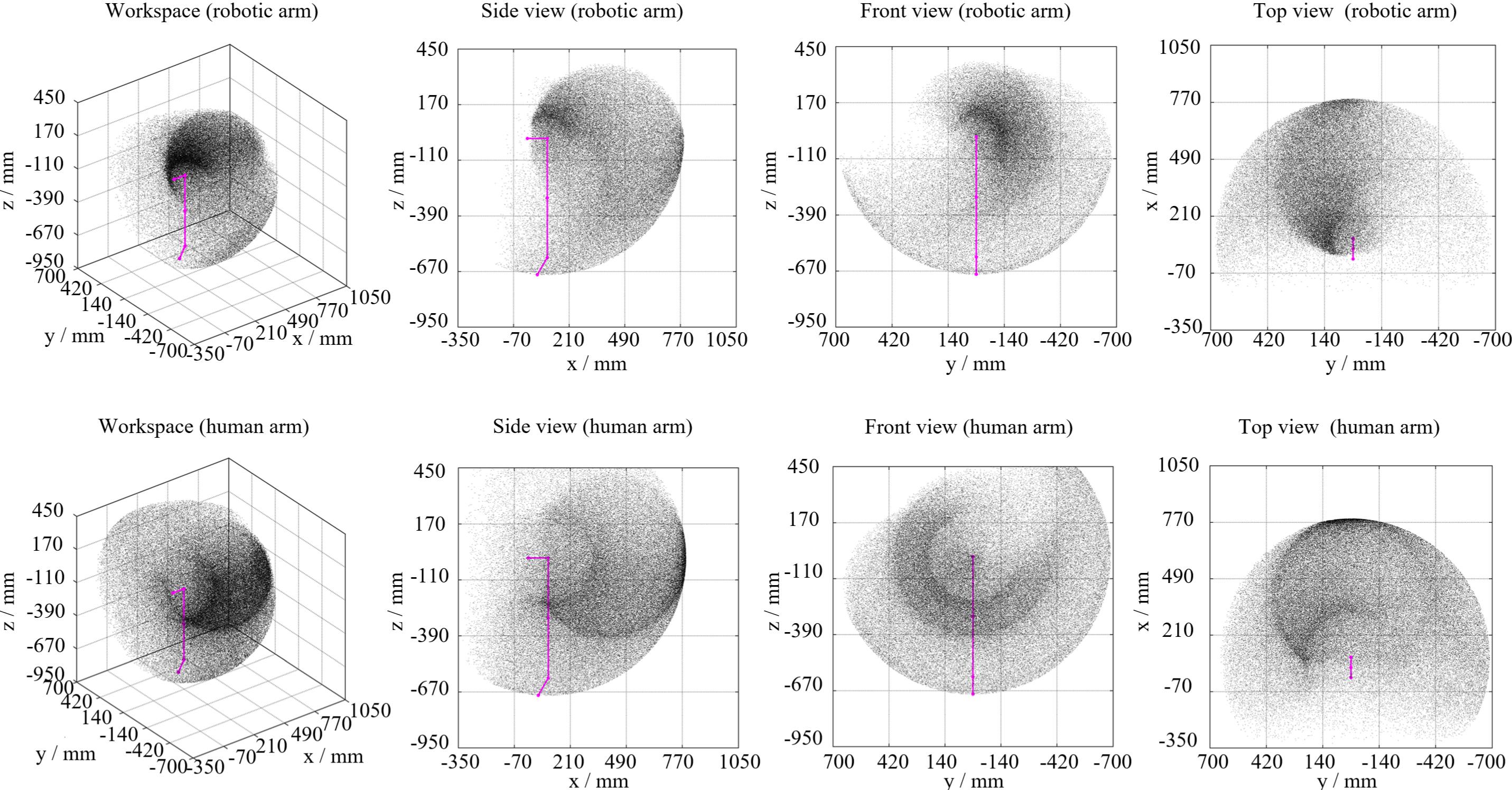}}
\caption{The workspace of the hand in the proposed robotic arm and human arm(Monte Carlo method is applied)}
\label{fig8.2}
\end{figure}

The range of motion for each joint is utilized to determine the workspace, or the reachable area of the robotic hand, as depicted in Fig. \ref{fig8.2}. This workspace closely mirrors that of the human arm (scapula motion is not considered) when maintaining the same limb length.

\subsection{Performance evaluation of the robotic arm powered by the compliant actuators.}

Owing to the high performance of the compliant actuators, the robotic arm exhibits versatility across various environments, despite its dimensions being constrained to human-arm proportions. This robotic arm can be used in a variety of scenarios, such as playing table tennis or badminton which require high end-effector speed, or handling heavy objects which demand high load capacity.

\begin{figure*}[htbp]
\centerline{\includegraphics[width=1\textwidth]{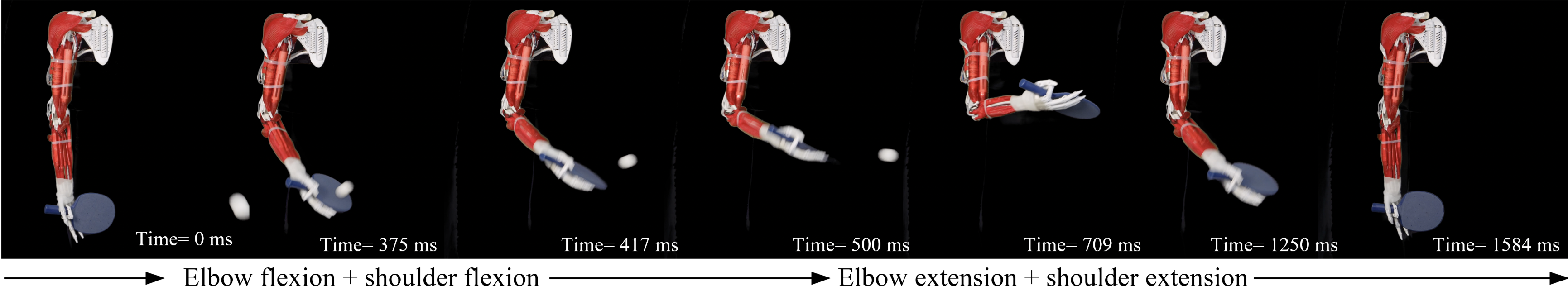}}
\caption{Table tennis playing test.}
\label{fig3.15}
\end{figure*}

Achieving high speed in industrial robotic arms is relatively straightforward and can typically be accomplished by incorporating appropriate motors and reducers based on design specifications. However, in the development of highly biomimetic robotic arms, stringent restrictions on size and form factor severely limit the available choices for drive mechanisms. Consequently, finding a solution ensuring ample output torque while achieving high-speed operation at the end-effector poses a substantial challenge. One example of a high-speed manipulator is the WAM Arm from Barret Technology, which weighs 27 kg and has a maximum end-effector speed of 3 m/s according to its official specifications\cite{senoo2006ball}. To test the performance of the compliant actuators in driving a robotic arm to output high speed, a table tennis-playing scenario was used. The arm was controlled so that the elbow and shoulder joints flexed simultaneously to hit the ping pong ball before returning to the initial position, as shown in Fig. \ref{fig3.15}. The test was conducted with a low load on the compliant actuators {(the ping pong paddle weighed 238 g, forearm weighed 1 kg, the whole arm weighted 4 kg)}, and the actuators could operate at their maximum speed (110 mm/s). The instantaneous speed of the end-effector reached 3.2 m/s, and the time taken from the start of the arm flexion to hitting the ping-pong ball was 188 ms.

\begin{figure*}[htbp]
\centerline{\includegraphics[width=1\textwidth]{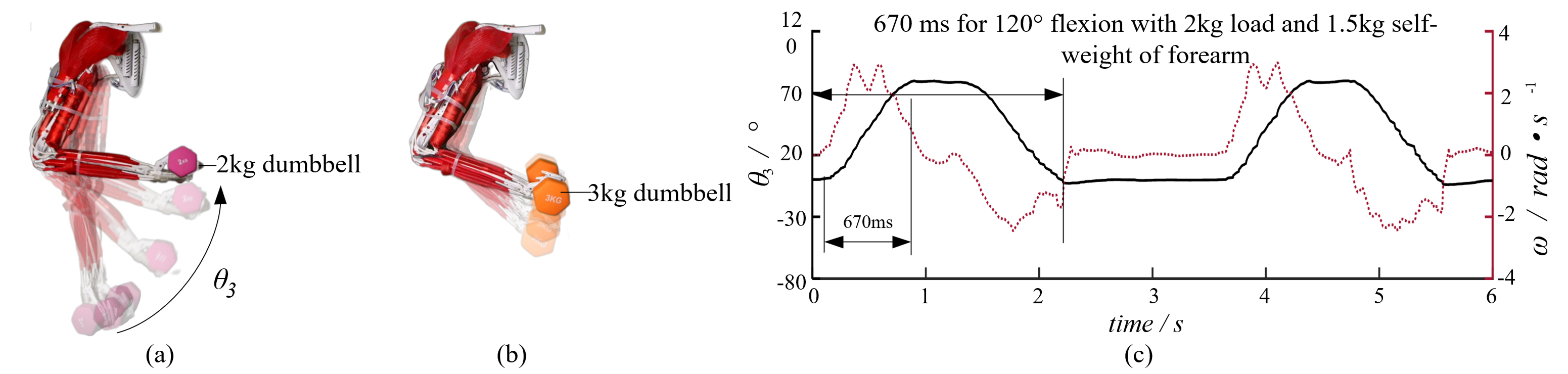}}
\caption{A dumbbell lifting test. (a) Test with 2kg dumbbell; (b) Test with 3kg dumbbell; (c) The forearm position $\theta_2$ and the angular velocity $\omega$ recorded at the maximum achievable speed as well as acceleration that did not overload the actuators.}
\label{fig3.12}
\end{figure*}

To demonstrate that the robotic arm with the proposed compliant actuators is capable of outputting a large payload at a fast speed, it imitates the human arm lifting dumbbells. The forearm's position was recorded using a gyroscope. In this bionic arm, elbow flexion can be driven by both the brachialis (MISA) and biceps (ECA).

The biomimetic robotic arm's performance was tested by holding a dumbbell (2 kg) with the elbow flexed to $120^{\circ}$ and then extended, as shown in Fig. \ref{fig3.12}(d). Two MISAs were used, and the motor was equipped with a safety mechanism that cuts off excessive speed or abrupt acceleration. {The associated test videos are provided in the supplemental materials.} The forearm's position $\theta_2$ and angular velocity $\omega$ were recorded, as shown in Fig. \ref{fig3.12}(e), with the maximum permitted speed and acceleration. The test results showed that the elbow fully flexed and lifted the 2 kg weight in 0.67 s, achieving over 0.74 Hz of maximum speed (excluding intervals in full flexion). At $\theta_3$ = 50\degree and $\omega$ = 3 rad/s, the maximum joint torque, including gravity resistance, surpassed 12 Nm, with the peak power registering at 36 W.

To validate the applicability of MISAs and ECAs to highly biomimetic robotic arms in performing high-intensity tasks, lifting tests were conducted with the robotic arm holding a 3kg dumbbell while both actuators worked simultaneously, as shown in Fig. \ref{fig3.12}(c). {However, the actuator's moment arm driving the joint is shorter when the elbow joint angle is either smaller, nearing full extension, or larger, approaching full flexion. In such cases, lifting a dumbbell necessitates a greater tension output from the actuator on the tendon. This phenomenon is mirrored in the human arm as well. To avert potential damage to the prototype, tests were deliberately not conducted at points of full extension and flexion of the elbow.} During the test, the peak joint output torque exceeded 16 Nm, taking into account the resistance of gravity, at lower joint speeds.

\begin{figure*}[htbp]
\centerline{\includegraphics[width=1\textwidth]{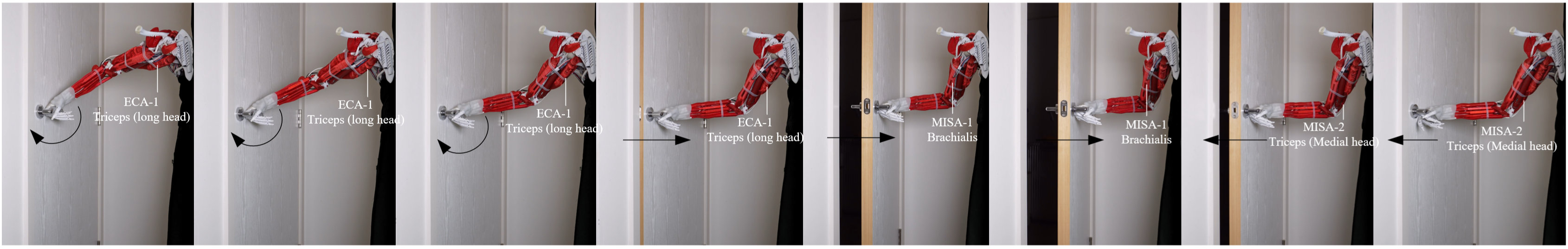}}
\caption{Door open test of a biomimetic robotic arm.}
\label{fig3.14}
\end{figure*}

Fig. \ref{fig3.14} shows the robotic arm is applied to the door opening task. In the test, the robotic hand is first actuated to hold the door handle by driving the ICA-1 to assist in forearm rotation. The long head of the triceps (ECA-1) then drives the shoulder adducted and turns the door handle to unlock the door. Next, the elbow was flexed by two MISAs and pulled the door. Then the elbow is extended and pushed the door back. {The associated test videos are provided in the supplemental materials.} Experimentally, a torque exceeding 1.5 Nm was required to unlock the door handle. This test may verify that the robotic arms with proposed compliant actuators can be used in challenging daily tasks.

{The list of videos for testing the proposed actuators and demonstrating the capabilities of the robotic arm are provided in Table \ref{tab12}. The supplementary video is accessible via the following link: \url{https://youtu.be/uunLkb-w0K0}.

\section{Discussions}

\begin{table}[htb]
\caption{Multimedia extensions are accessible via the link: \url{https://youtu.be/uunLkb-w0K0}}
\footnotesize
\begin{center}
\begin{tabular}{c l}
\toprule
No.   & Description  \\
\midrule
Video 1 & Force-displacement relationship test \\
Video 2 & Experimental of joint stiffness adjustment by applying two MISAs \\
Video 3 & Accomplishment of variable joint stiffness with MISAs - passive mode \\
Video 4 & Accomplishment of variable joint stiffness with MISAs - active mod \\
Video 5 & Passive performance of the robotic arm by applying compliant actuators \\
Video 6 & Table tennis playing test \\
Video 7 & Dumbell lifting test (2kg and 3kg) \\
Video 8 & Door open test \\ 
\bottomrule
\end{tabular}
\label{tab12}
\end{center}
\end{table}

The results of the table tennis playing and dumbbell lifting tests provided evidence of the high power output of MISAs, despite their small size and weight. The door opening test confirms the versatility of the proposed three actuators and their ability to work together seamlessly to execute complex actions. The successful completion of these tasks underscores the potential of highly biomimetic robotic arms equipped with MISAs, ICAs, and ECAs to perform challenging tasks with high-intensity demands without sacrificing the natural appearance of human-like robots. Overall, these experiments demonstrate the great potential of proposed actuators in enabling highly functional and aesthetically pleasing robots that can perform a wide range of tasks with exceptional power.}

These three motor-based artificial muscles are intended for highly biomimetic robotic arms. This is a new direction in robotics design: all robot joints without rigid shafts, using ligaments to stabilise the joints; driven by muscles that are not mechanically fixed and have similar human arm muscle arrangement, imitating the actuation method used in human joints; a skeletal model of the human arm is used directly as the main structure of the robot. Unlike conventional robotic arms where the joints are driven directly by motors, the actuators of the biomimetic robotic arm are placed parallel to the skeleton, achieving compactness with high space utilisation and allowing the use of high-power actuators. From the simulations and experimental results, it was preliminarily evidenced that MISA had the highest performance. The advantage of MISA, which was described in \cite{yang2023novel}, enables a wide range of joint stiffness, and achieves a high power-to-volume ratio (345$\times$10$^{3}$ W/m$^{3}$). But each of the three actuators had irreplaceable advantages and was therefore applied to the robotic arm prototype.

For the ICA, the unique advantage is that it can be applied in both remote and local tendon-driven methods. When the ICA works in unidirectional output mode, the actuator can be mechanically fixed and the contracted tendon can drive the joint remotely. Both ECA and MISA can only be used in local tendon-driven, where two tendons are connected to the driving arm and driven arm respectively. MISA and ECA require axial movement and sufficient installation space. In the development of the robotic arm prototype, a 22 mm diameter motor was used to achieve a large torque of forearm pronation. Due to limited installation space, {Implementing MISA or ECA at the location of the pronator teres proves to be impractical with the local tendon-driven approach.} Thus, the ICA was mechanically fixed to the humerus to drive the forearm pronation with the remote tendon-driven method.

The first unique advantage of the ECA is its short length. The total actuator length is only 7.2\% longer compared to the motor, whereas it is 35.8\% for the ICA and 59.1\% for the MISA. The second unique advantage is the selection of springs that allow for a high coefficient of elasticity. High-torque motors can be used to achieve large tendon output forces. For the MISA, the maximum repulsive force between the magnets limits the maximum tendon output force (excessive force would disable the protection, and large repulsive force requires oversized magnets). As for the ECA, it allows for a large spring outer diameter with a large maximum compression force, and a large maximum tendon output force can be achieved. In the robotic arm prototype, the elbow joint is driven by the brachialis, biceps, and triceps. The brachialis and triceps can be antagonistically arranged (both connected to the ulna) using a pair of MISAs, which allows variable joint stiffness. The biceps have no antagonistic muscles attached to the radius in the robotic arm prototype, so even if the MISA is used as the biceps, the variable stiffness function of the humeroradial joint will not be realized. The MISA is not considered to function as the biceps muscle. In addition to actuating the forearm supination, the biceps also assist in driving the elbow joint flexion, so that the large tendon output force can significantly increase the joint output torque. Based on the considerations listed above, the ECA was used to simulate the biceps by combining the drawbacks of the MISA with the two unique advantages of the ECA.

\begin{figure*}[htbp]
\centerline{\includegraphics[width=1\textwidth]{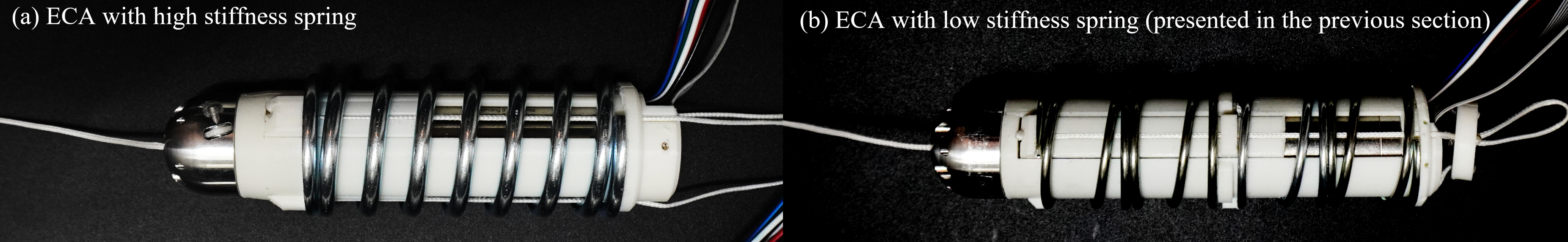}}
\caption{(a) ECA with high stiffness spring; (b) Previous discussed ECA.}
\label{fig3.121}
\end{figure*}

For the Deltoid muscles, ensuring robust output force is pivotal for delivering adequate torque to the shoulder joint. An inadequately elastic spring risks reaching its solid state even under no load, rendering the elastic component of the actuator ineffective upon load application. Among various actuators evaluated, only the ECA can be adapted with a high-stiffness elastic element and possesses a straightforward adjustment for peak output force. Consequently, the ECA, equipped with a higher stiffness spring, was selected as the actuator for the deltoid. As illustrated in Fig. \ref{fig3.121}(a), this ECA features a compression spring with a stiffness of 14.5N/mm and a peak elastic force of 622.7N, making it suitable as a Deltoid actuator. This configuration, when compared to the ECA with a lower stiffness spring depicted in Fig. \ref{fig3.121}(b) from prior analyses, has a larger diameter and increased weight. However, its elastic force and output forces are augmented by 2.5 times due to the incorporation of a more potent motor (Maxon EC-4 pole, 22mm, 90 W).

The comparison of different compact SEAs is shown in Table \ref{tab9}. In the ECA design (previously analysed), the spring is sleeved over the motor. This design maximises space utilisation and therefore has a significantly higher power-to-volume ratio, which is 361$\times$10$^{3}$W/m$^{3}$. the ICA uses a lighter torsion spring and therefore has a higher power-to-mass ratio than the SEAs including the ECA, which is 111.6 W/kg.

\begin{table*}[htbp]
\caption{Comparison of different compact SEAs}\label{tab9}
\scriptsize
\centering
\begin{threeparttable} 
\begin{tabular}{l l l l l l l l l l l }
\toprule
 SEA & Type$^A$ & Output$^B$ & Elastic$^C$ & Power         & Weight  & Volume                & Power/Volume        & Power/Mass & Bio$^D$ \\
      &      & (Nm or N)      &(mm or \degree)       & (W)    & (kg)  & ($\times$10$^{-6}$m$^3$) &($\times$10$^{3}$W/m$^3$)& (W/kg)     &  \\
\midrule
UT-SEA\cite{paine2013design}  & Linear & 848 N & 60mm  & 110 & 1.168  & / & / & 94 & No \\

SAFFiR \cite{lee2013design}  & Linear & 300 N & 110mm & \textless 80$^E$ & 0.816  & / & / & \textless 98 & No \\

THOR\cite{knabe2014design} & Linear & 600 N & 85mm  & \textless 80$^E$ & 0.938 & / & / & \textless 85 & No \\

SEA\cite{pratt2002series} & Linear & 136 N  & / & 44.76 & 0.4536 & 333 & 134 & 98.7 & No \\

cRSEA\cite{kong2011compact} & Rotary & 10 Nm & $\pm$50 \degree  & \textless 135$^E$ & / & / & / & / & No \\

SEA\cite{sergi2012design} & Rotary & 10 Nm & $\pm$6 \degree  & \textless 65$^E$ & 1.8 & 1865 & 34.85 & \textless 36.1 & Yes \\

HT-SEA\cite{agarwal2017series} & Tendon-R & 0.3 Nm & / & 13.86$^D$ & 0.393$^F$ & 100.131$^F$ & \textless 138 & \textless 35.26 & Yes \\

LC-SEA\cite{agarwal2015index} & Tendon-R & 0.5 Nm & /  & 13.86$^D$ & 0.383$^F$ & 92.403$^F$ & \textless 149 & \textless 36.03 & Yes \\

\textbf{ICA (proposed)}  & Tendon-L & 125 N & 34.8 mm & 31.248 & 0.28 & 90.78 & 344 & 111.6 & Yes \\

\textbf{ECA (proposed)}  & Tendon-L & 250 N & 28.5 mm  & 31.248 & 0.295 & 86.53 & 361 & 105.9 & Yes \\

\bottomrule
\end{tabular}
\begin{tablenotes}
\footnotesize
\item[]$^A$ The tendon-driven features two modes: R (remote-control) and L (local-control). $^B$ Maximum output torque or force. $^C$Range of passive stroke or passive angular deflection. $^D$Whether can be applied to highly biomimetic robotics. $^E$ Not given in the literature, estimated according to the power of the motor used. $^F$ The data given in the literature are excluding the motor, they are the estimated value by adding the motor's weight or volume.
\end{tablenotes}
\end{threeparttable} 
\end{table*}

\section{Conclusion}

This paper delineates the design and assessment of two novel compliant actuators, the Internal Torsion Spring Compliant Actuator (ICA) and the External Spring Compliant Actuator (ECA), in juxtaposition with the formerly introduced Magnet Integrated Soft Actuator (MISA). The actuation mechanisms were elucidated, prototypes contrived, and their performance meticulously evaluated. All three actuators manifest aptitude for human-robot interaction and highly biomimetic robot applications, showcasing versatility across various robotic joint configurations without necessitating modifications. This facilitates enhanced naturalism and power output in compact limb designs as opposed to the conventional direct joint-motor shaft linkage. The localized placement of these proposed actuators near the joints augments their functional resemblance to artificial muscles, marking an advancement over existing tendon-driven SEAs.

ICAs are particularly suited for joints with spatial constraints yet demanding high-speed functionality, exhibiting a notable power-to-mass ratio of 111.6 W/kg. On the other hand, ECAs are conducive for joints necessitating substantial load capacities within confined spaces, with an impressive power-to-volume ratio of 361 x 10³ W/m³. They also permit the utilization of compression springs with diverse elastic properties, ensuring remarkable load bearing without elongating the actuator. Conversely, MISAs facilitate a broad spectrum of controllable stiffness in joints, a feature notably superior to that of ECAs.

The furtherance of this research through rigorous testing—encompassing high-intensity dumbbell lifting, table tennis playing, and door opening—affirms the operational efficacy and visual human resemblance of the proposed local tendon-driven actuators amalgamating both SEA and VSA designs. The success in executing complex tasks underlines the potential of these innovative compliant actuators in fostering the development of highly biomimetic robotic systems, thereby paving the way for more natural and intuitive human-robot interactions.

\newpage

\bibliographystyle{SageV}
\bibliography{Reference.bib}



\end{document}